\documentclass{article}

\usepackage{microtype}
\usepackage{graphicx}
\usepackage{subcaption}
\usepackage{booktabs} % for professional tables
\usepackage{multirow}
\usepackage{multicol}
\usepackage{xcolor}         % colors
\usepackage[table]{xcolor}
\usepackage{wrapfig}
\usepackage{makecell}
\usepackage{etoc}
\usepackage{colortbl}
\usepackage{etoolbox}
\definecolor{VintageBlueAAA}{HTML}{F7FCF0}
\definecolor{VintageBlueAA}{HTML}{E0F3DB}
\definecolor{VintageBlueA}{HTML}{CCEBC5}
\definecolor{VintageBlueB}{HTML}{A8DDB5}
\definecolor{VintageBlueC}{HTML}{7BCCC4}
\definecolor{VintageBlueD}{HTML}{4EB3D3}
\definecolor{VintageBlueE}{HTML}{2B8CBE}
\definecolor{VintageBlueF}{HTML}{0868AC}
\definecolor{VintageBlueG}{HTML}{084081}

\definecolor{CreamAAA}{HTML}{FCFBFD}
\definecolor{CreamAA}{HTML}{EFEDF5}
\definecolor{CreamA}{HTML}{DADAEB}
\definecolor{CreamB}{HTML}{BCBDDC}
\definecolor{CreamC}{HTML}{9E9AC8}
\definecolor{CreamD}{HTML}{807DBA}
\definecolor{CreamE}{HTML}{6A51A3}
\definecolor{CreamF}{HTML}{54278F}
\definecolor{CreamG}{HTML}{3F007D}

\definecolor{LightGray}{HTML}{ECECEC} 
\definecolor{VintagePinkE}{HTML}{D8758F} 

\newcommand{\ColorSquare}[1]{%
  \textcolor{#1}{\rule{1em}{0.8em}}%
}

\usepackage{calc}
\usepackage{tcolorbox}

\newlength{\CardW}
\newlength{\CardH}
\newlength{\CapH}
\newlength{\ImgH}
\usepackage{hyperref}
\usepackage[accepted]{icml2026}

\usepackage{amsmath}
\usepackage{amssymb}
\usepackage{mathtools}
\usepackage{amsthm}

\usepackage[capitalize,noabbrev]{cleveref}

\theoremstyle{plain}

\theoremstyle{definition}

\theoremstyle{remark}

\usepackage[normalem]{ulem}

\usepackage[textsize=tiny]{todonotes}

\icmltitlerunning{Next-Gen CAPTCHAs}

\begin{document}

\twocolumn[
  \icmltitle{Next-Gen CAPTCHAs: Leveraging the Cognitive Gap for Scalable and Diverse GUI-Agent Defense}



  \icmlsetsymbol{equal}{*}

  \begin{icmlauthorlist}
    \icmlauthor{Jiacheng Liu}{equal,mbz,meta}
    \icmlauthor{Yaxin Luo}{equal,mbz,meta}
    \icmlauthor{Jiacheng Cui}{mbz,meta}
    \icmlauthor{Xinyi Shang}{mbz,meta,ucl}
    \icmlauthor{Xiaohan Zhao}{mbz,meta}
    \icmlauthor{Zhiqiang Shen}{mbz,meta}
  \end{icmlauthorlist}

  \icmlaffiliation{mbz}{Mohamed bin Zayed University of Artificial Intelligence (MBZUAI)}
  \icmlaffiliation{meta}{MetaAgentX}
  \icmlaffiliation{ucl}{University College London}

  \icmlcorrespondingauthor{Zhiqiang Shen}{Zhiqiang.shen@mbzuai.ac.ae}

  \icmlkeywords{Agent Security, CAPTCHAs, GUI/Computer Use Agent, Web Agent}

  \vskip 0.3in
]



\printAffiliationsAndNotice{\icmlEqualContribution}

\begin{abstract}
The rapid evolution of GUI-enabled agents has rendered traditional CAPTCHAs obsolete. While previous benchmarks like OpenCaptchaWorld established a baseline for evaluating multimodal agents, recent advancements in reasoning-heavy models, such as Gemini3-Pro-High and GPT-5.2-Xhigh have effectively collapsed this security barrier, achieving pass rates as high as 90\% on complex logic puzzles like ``Bingo''. In response, we introduce Next-Gen CAPTCHAs, a scalable defense framework designed to secure the next-generation web against  the advanced agents. Unlike static datasets, our benchmark is built upon a robust data generation pipeline, allowing for large-scale and easily scalable evaluations, notably, for backend-supported types, our system is capable of generating effectively unbounded CAPTCHA instances. We exploit the persistent human--agent ``Cognitive Gap'' in interactive perception, memory, decision-making, and action. By engineering dynamic tasks that require adaptive intuition rather than granular planning, we re-establish a robust distinction between biological users and artificial agents, offering a scalable and diverse defense mechanism for the agentic era. \footnote{Project page is available at \url{https://greenoso.github.io/NextGen-CAPTCHAs_webpage/}.}
\end{abstract}

\section{Introduction}

CAPTCHAs (Completely Automated Public Turing test to tell Computers and Humans Apart)~\citep{vonAhn2003captcha} have long served as a lightweight, widely deployed line of defense for the open web, limiting automated abuse such as credential stuffing, scraping, fake account creation, and transaction fraud. Historically, CAPTCHA design has followed an arms race~\citep{captcha_security,endisnigh,arm_race}: early distorted-text challenges aimed to resist OCR, later image-grid CAPTCHAs targeted object recognition, and more recent “logic” CAPTCHAs introduced game-like spatial reasoning. In each cycle, progress in machine perception and reasoning steadily eroded the security margin of what was once considered ``bot-hard'', forcing defenders to continuously redesign challenges and patch deployed systems.

\begin{figure}[t]
\centering
\includegraphics[width=1.0\linewidth]{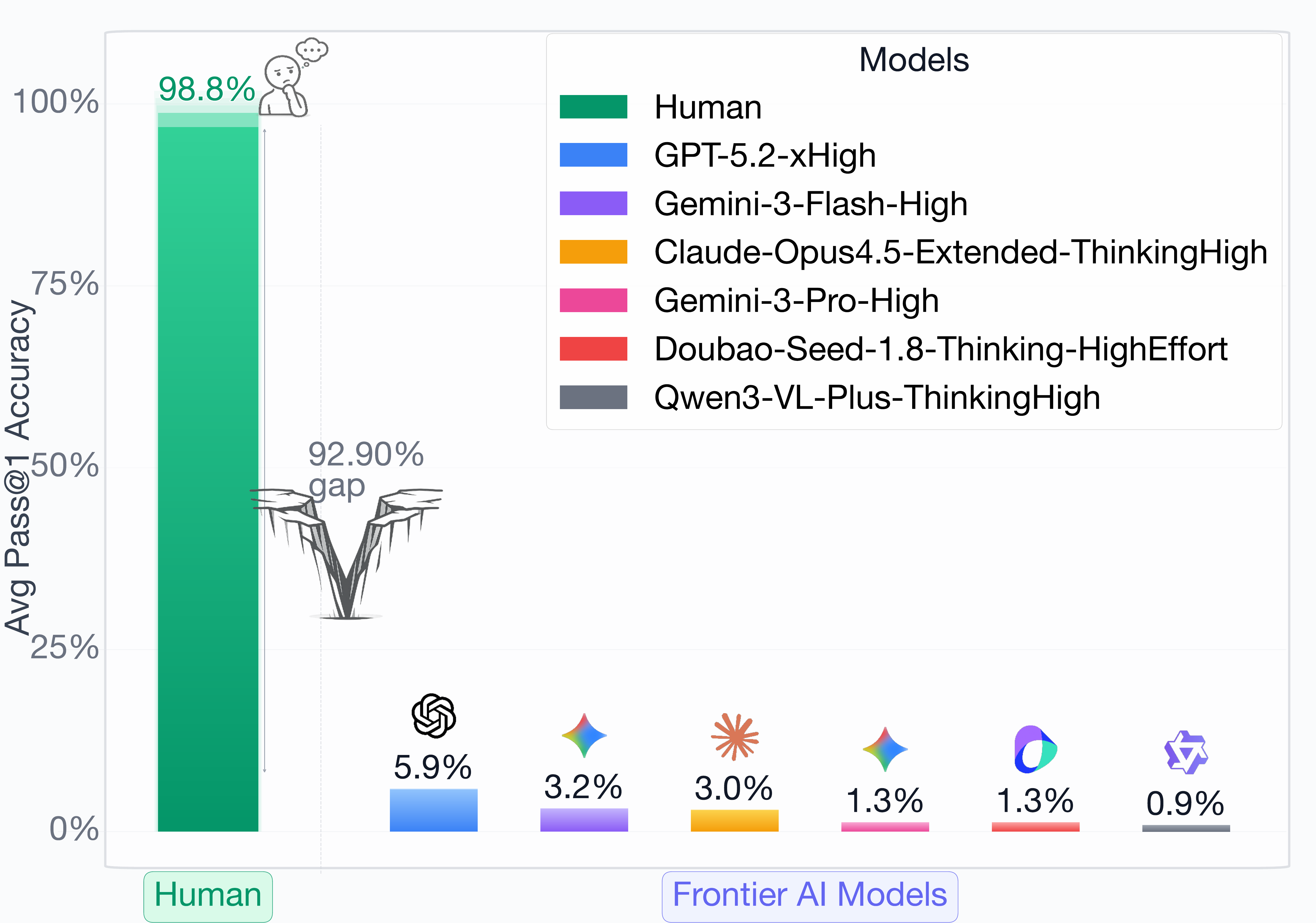}  
\vspace{-1.5em}  
\caption{Frontier Models as GUI Agent Backbones' Pass@1 on our Next-Gen CAPTCHA benchmark.} 
\label{benchmark_main}
\vspace{-0.25in}
\end{figure}

This erosion has accelerated sharply with the emergence of Multimodal Large Language Models (MLLMs) and \emph{GUI-enabled / Computer-Use} agents. Modern web agents can perceive rendered pages, interpret instructions, and execute multi-step interactions—capabilities demonstrated on realistic benchmarks such as Mind2Web and WebArena~\citep{deng2023mind2web,zhou2023webarena}. Recent security analyses further indicate that these agentic capabilities generalize zero-shot to diverse CAPTCHA challenges, effectively neutralizing the long-standing bot-hard assumption that underpins modern web security~\citep{teoh2025captchas}. The risk is no longer theoretical: as computer-use agents become integrated into consumer and enterprise products, attackers gain an increasingly accessible “automation substrate” that can operate directly through the same browser surfaces intended for humans, widening the attack surface of websites that rely on current CAPTCHAs as a primary gatekeeper.

At the same time, substantial capability gaps remain between humans and current MLLM-based agents, especially in \emph{interactive} settings. Beyond static vision--language benchmarks, agents must repeatedly (i) ground instructions to precise screen regions, (ii) maintain and update a latent task state over time, and (iii) execute low-level actions robustly under partial observability and UI stochasticity. Prior work reports persistent deficits in visual--spatial grounding and intermediate state manipulation in MLLMs, and shows that offline benchmark success often degrades when models are deployed as interactive web agents in live environments~\citep{cao2024visual,yang2025thinking,xue2025illusion}. These gaps suggest an opportunity for defense: rather than “hardening” existing decomposable puzzles, we can design challenges that are trivial for human intuition but systematically misaligned with the over-segmented, step-by-step strategies used by today’s agents.

Motivated by this, we introduce Next-Gen CAPTCHAs: a scalable GUI-agent defense framework that produces GUI agent-defensive yet human-solvable interactive challenges. The core of Next-Gen is an automatic CAPTCHA generation pipeline with rule-based verifiable answer: for generative task families, the system can synthesize an effectively unbounded number of unique instances (e.g., timed interaction tasks such as \texttt{Red Dot}), each paired with a rule-based solution—eliminating the need for human annotation while maintaining strong instance diversity. Importantly, our \textbf{27 types of CAPTCHA families are newly designed by us} specifically to defend against modern GUI agents (rather than collected from existing deployed CAPTCHA systems). Except for two vision--language families, instances are procedurally generated with automatically verifiable solutions.

To make progress measurable and reproducible, we curate a \emph{benchmark sample} \emph{from} this defense system: a main test set of 519 vision--language puzzles spanning the 27 vision--language families, plus a lightweight subset with 5 puzzles per task for cost-effective evaluation under limited budgets. Under realistic closed-API agent settings on live web tasks, we observe a large human--agent gap: humans achieve near-ceiling solve rates with low latency, whereas representative high-reasoning MLLMs and GUI web agents exhibit single-digit Pass@1 (Fig.~\ref{benchmark_main}). We further validate human friendliness via a small-scale human study, reporting high success rates and low completion times across representative tasks.

Finally, we release a real-web evaluation platform that is agnostic of GUI framework: any GUI-enabled MLLM agent can be evaluated via a standardized browser interaction and logging interface. In our experiments, we use \texttt{Browser-Use} as the primary reference integration, and additionally evaluate supplementary agent frameworks including \texttt{Claude Cowork} and \texttt{CrewAI}. While the curated benchmark facilitates standardized comparison, it is best viewed as a \emph{byproduct} of the broader Next-Gen defense system: the dataset is sampled for benchmarking, whereas the primary goal is a deployable, continuously generative CAPTCHA defense mechanism for the agentic era.

Our contributions are threefold:
(1) We design 27 \emph{new} formats of Next-Gen CAPTCHA families that target empirically observed human--MLLM gaps, aiming to be agent-defensive while remaining human-friendly.
(2) We propose a procedural generation and automatic verification framework, enabling scalable deployment with effective diversity and controllable difficulty.
(3) We release an open-source real-web evaluation platform and a benchmark \emph{sampled from the defense system}, including both a 519-puzzle main set and a cost-aware subset, and evaluate GUI Agents across perception, memory, reasoning, and action under realistic interaction constraints. 
\section{Background}

CAPTCHA systems have long been shaped by an \textit{arms race} with AI~\citep{vonAhn2003captcha,endisnigh}. We summarize this progression across three capability shifts: \textit{Perception}, \textit{Reasoning}, and \textit{Agency}, where defenses were eventually broken by the very models they aimed to stop.

\noindent\textbf{The Era of Visual Perception.} The defense against automated bots initially relied on distorted text to defeat OCR, premised on the belief that decoding warped visuals was uniquely human~\citep{vonAhn2003captcha, shet2014you}. However, Convolutional Neural Networks (CNNs) rendered this obsolete by solving text challenges with superhuman accuracy~\citep{mori2003breakingCaptcha, shet2014you,gao2016simpleGenericAttack}. Consequently, security mechanisms shifted to object classification~\citep{googleRecaptchaV2DocsDisplay}, assuming machines lacked the contextual grounding to identify objects in diverse scenes. Yet, the emergence of Vision Transformers~\citep{dosovitskiy2021vit} and large-scale pre-training has since bridged this gap. Modern backbones can now interpret complex scenes with precision~\citep{plesner2024breaking, sivakorn2016iamrobot,hossen2020object}, effectively neutralizing static visual perception as a reliable security barrier.

\noindent\textbf{The Era of Multimodal Logic: Reasoning vs. MLLMs.} To counter advanced vision models, providers like Arkose Labs introduced ``Logic CAPTCHAs"~\citep{arkoseMatchKey} puzzles requiring not just recognition, but spatial reasoning and game-like logic (e.g., rotating objects or matching icons). Until recently, these were considered secure against standard 
vision models. However, the emergence of Multimodal Large Language Models (MLLMs) like GPT-5.2~\citep{gpt5.2}, Claude4.5-Opus~\citep{claude4.5} and Gemini3-Pro~\citep{gemini3} changed the landscape. Recent benchmarks, such as MCA-Bench~\citep{wu2025mca} and COGNITION~\citep{wang2025cognition}, revealed that MLLMs could effectively interpret instructions and solve these logic puzzles. Crucially, as reasoning-enhanced models continue to scale, even complex logic tasks like ``Bingo'' is seeing pass rates exceed 90\%. In a security context, a pass rate of even 50\% is a critical failure; for robust defense, machine success rates must be driven down to near 0\%, as a single successful breach allows an agent to access sensitive data.

\begin{figure}[t]
\centering
\includegraphics[width=1.0\linewidth]{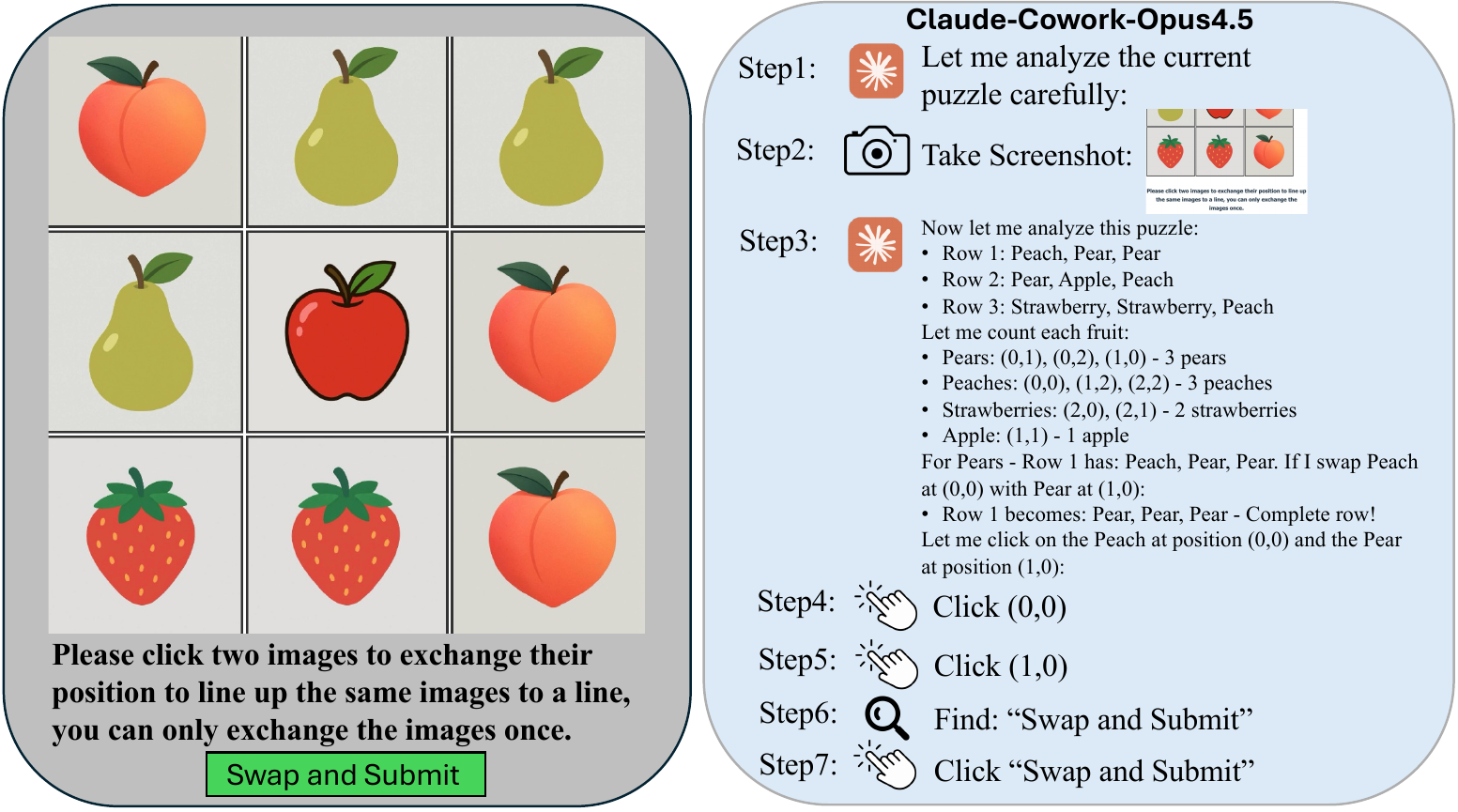}  
\vspace{-1.3em}  
\caption{With the enhanced Computer Use abilities like taking screenshot, clicking and dragging etc, along with advanced thinking and tool-use power (e.g, searching), Claude-Cowork-Opus4.5 can now solve ``Bingo'' CAPTCHA effectively and efficiently.}
\vspace{-1.5em}  
\label{section3_case_analysis}
\end{figure}

\noindent\textbf{The Era of Agentic AI: The Unaddressed Threat of ``GUI Agents''.} We are now entering a fourth, unprecedented era: the age of Autonomous Web Agents equipped with Computer Use and GUI interaction capabilities~\citep{anthropic2024computerUse,openai2025chatgptAgent,claudeCowork}. Unlike passive solvers, these agents can plan, navigate, and execute multi-step workflows~\citep{xue2025illusion}. Existing literature primarily focused on evaluating agent capability rather than defending against it~\citep{luo2025open,bhardwaj2026beyond,zhang2025capture,kharlamova2025spatial}. Although some works attempt to design individual captchas~\citep{ding2025illusioncaptcha}, there are currently no systematic defenses designed specifically to prevent these agents. Current CAPTCHAs fail to differentiate between a human user and an agent that can ``think'' through a workflow~\citep{deng2025oedipus,qi2026viper}. This creates a critical vulnerability: existing defenses are now merely speed bumps for agents that can reason as effectively as humans~\citep{teoh2025captchas}.

\begin{figure}[t]
\centering
\vspace{-0.06in}
\includegraphics[width=1.0\linewidth]{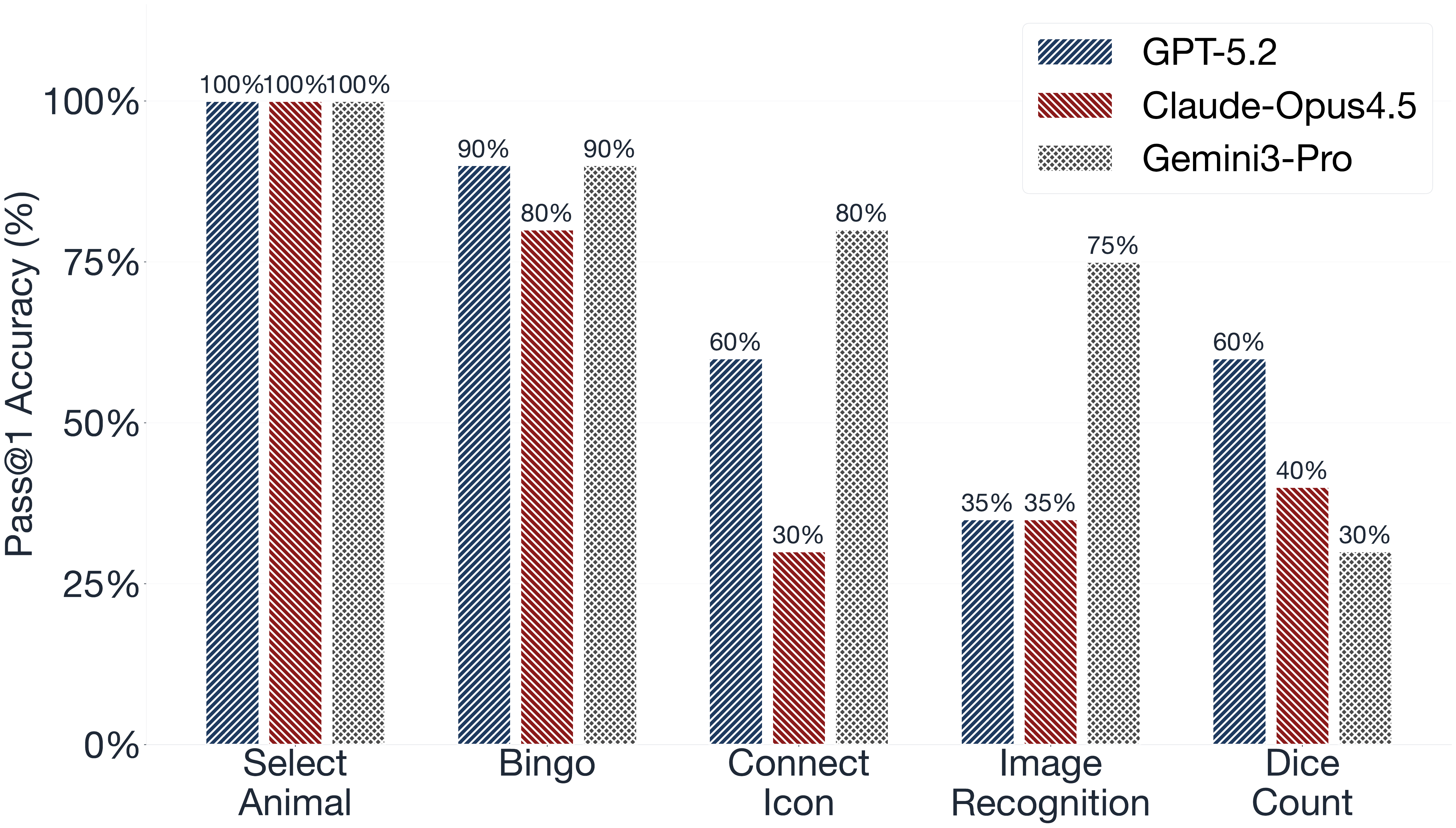}  
\vspace{-1.1em}  
\caption{\textbf{Current CAPTCHAs System Fails.} Recent Advanced MLLMs can even break the current CAPTCHAs system without extra high thinking efforts (We use the default thinking settings from their official API)} 
\label{captcha_breaks}
\vspace{-0.16in}
\end{figure}

This paper addresses this specific void. We argue that defending against the next generation of agents requires moving beyond static difficulty to exploiting the ``Cognitive Gap", designing tasks that are trivial for human intuition but computationally exhaustive for the over-planning nature of agentic reasoning.

\vspace{-0.05in}
\section{Advanced GUI Agents Break Current CAPTCHA System}
\label{section3}
As shown in Fig.~\ref{captcha_breaks}, the security barrier provided by many current CAPTCHA systems has effectively collapsed under state-of-the-art MLLMs powered GUI Agents. On traditional logic puzzles (e.g., \textit{Select Animal}), leading models such as Gemini3-Pro and GPT-5.2 reach 100\% Pass@1. Even on more complex reasoning tasks like \textit{Bingo}, which requires cross-referencing visual evidence with symbolic rules, models achieve 80--90\% Pass@1. Overall, \emph{static logic is no longer a bot-hard problem}.

\begin{figure*}[!h]
\centering
\includegraphics[width=1.0\linewidth]{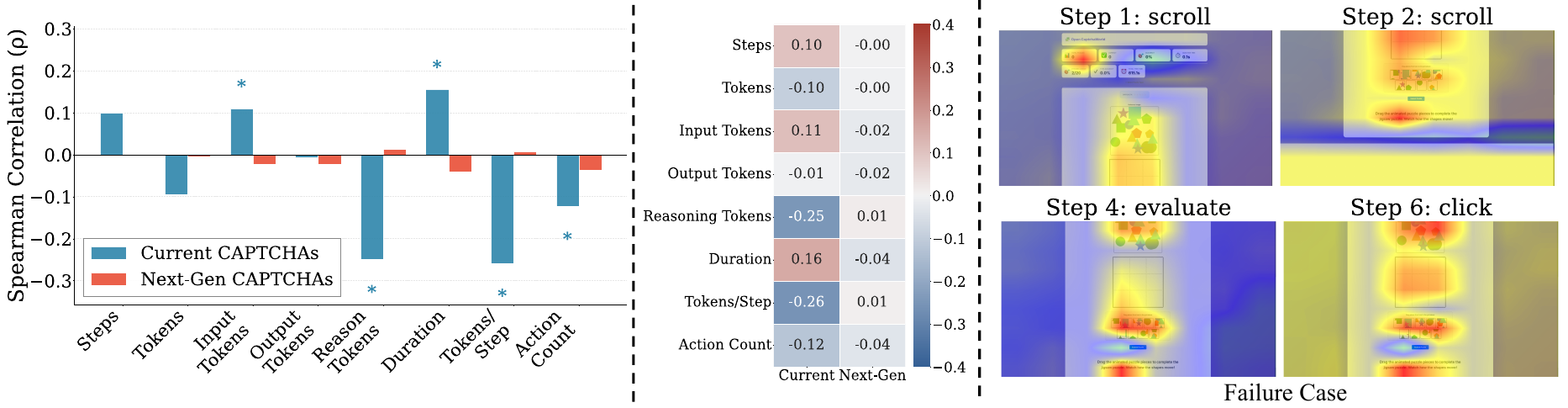}  
\vspace{-1.1em}  
\caption{\textbf{Success--trajectory correlation differs between current and Next-Gen CAPTCHAs.}
\textbf{Left \& Middle}: Spearman $\rho$ between each CAPTCHA family's Pass@1 and logged trajectory metrics (bar plot on the left; heatmap on the right; * denotes significance).
Current CAPTCHAs show non-trivial correlations, while Next-Gen correlations are near zero.
\textbf{Right}: a representative Next-Gen failure on a jigsaw-like puzzle, where the agent scrolls/evaluates but skips drag-and-drop and prematurely clicks \texttt{Submit}.}
\vspace{-1em}  
\label{correlation_analysis}
\end{figure*}
\noindent\textbf{Perception--Thinking--Action Analysis of GUI Agents.}
To understand why advanced GUI agents solve prior CAPTCHA designs so effectively, we analyze Claude-Cowork-Opus4.5 (Fig.~\ref{section3_case_analysis}). The core issue is structural: most current CAPTCHAs are short, decomposable workflows. Once agent obtains an observation of the page state, the remaining problem becomes a sequence of locally verifiable micro-decisions followed by straightforward UI operations, a regime where modern tool-using agents are highly reliable.
We illustrate this with the \textit{Bingo} CAPTCHA in Fig.~\ref{section3_case_analysis}. The agent (i) screenshots and extracts a structured grid state, (ii) enumerates candidate swaps under the rule, and (iii) executes two clicks and submits. A representative fragment is:
Take Screenshot $\rightarrow$ parse grid into structured state (``Row~1: \dots'') $\rightarrow$
Find a swap that completes a row $\rightarrow$ click two cells $\rightarrow$ ``Swap and Submit''

This highlights a key point: once a CAPTCHA reduces to a small search over a discrete state space (grid states, icon matches, counting outcomes, or arithmetic choices), an agent with accurate perception and reliable interaction can solve it consistently because each step can be checked locally and the interface provides immediate feedback.

\noindent\textbf{Correlation Analysis Reveals a Structural Difference from Current CAPTCHAs.}
Beyond qualitative traces, we quantify how success relates to trajectory behavior by computing correlations between each CAPTCHA family's Pass@1 and logged interaction/reasoning metrics. Fig.~\ref{correlation_analysis} contrasts widely used web-service CAPTCHAs~\citep{luo2025open} (e.g., Bingo, Dice Count) with our Next-Gen families, evaluated using a Browser-Use agent powered by Gemini3-Flash-High.
For current CAPTCHAs, Pass@1 shows non-trivial correlations: success is weakly positively correlated with interaction length (e.g., Steps, Duration), while being negatively correlated with reasoning-heavy signals (e.g., Reasoning Tokens, Tokens-per-Step). This pattern is consistent with existing CAPTCHAs being \emph{workflow-friendly}: once the page state is observed, solving reduces to short, decomposable micro-decisions and UI actions, and extra deliberation often reflects uncertainty rather than progress.

For Next-Gen CAPTCHAs, correlations stay near zero across metrics. More steps or a larger deliberation budget rarely improve success. Fig.~\ref{correlation_analysis} (right) shows why: the agent may attend to the right region and still choose the wrong \emph{action primitive}. In the jigsaw-like CAPTCHA, solving requires drag-and-drop. The agent never drags and clicks \texttt{Submit} at Step~6. These failures come from interactive bottlenecks: misreading UI affordances, weak grounding for manipulation targets, or premature termination. Extra textual reasoning does not fix them.
Taken together, stronger multimodal perception plus computer-use already breaks many current CAPTCHAs by following decomposable workflows. We therefore avoid making the same tasks ``harder''. Instead, we design CAPTCHAs that exploit the Cognitive Gap: they require robust interactive grounding and manipulation under realistic interaction constraints.

\section{Next-Gen CAPTCHAs: Native GUI-Agent Era's Security Defense}

\begin{figure*}[t]
  \centering
  \includegraphics[width=\textwidth]{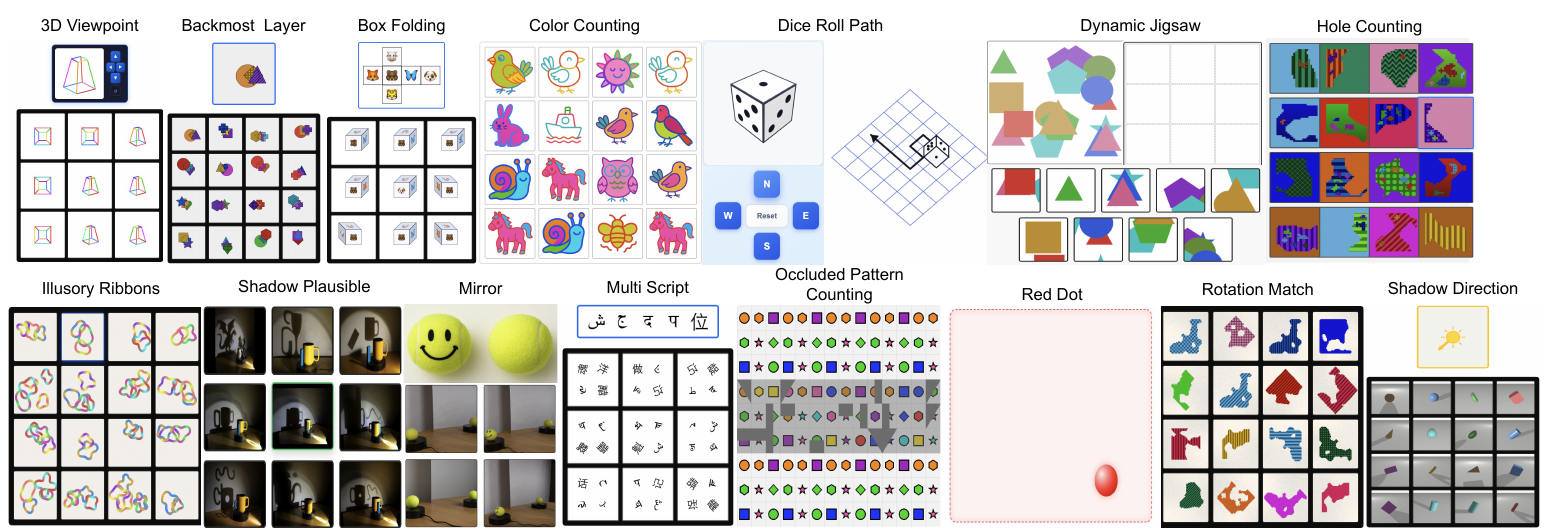}
  \vspace{-0.6em}
  \caption{\textbf{Next-Gen CAPTCHA family examples.}
  Representative instances from Next-Gen CAPTCHA task families.
The full family list is in Table~\ref{tab:captcha_families}; additional examples for all families are in Appendix~\ref{appendix:captcha_gallery}.}
  \label{fig:nextgen_examples}
  \vspace{-0.3em}
\end{figure*}

\subsection{Agent-CAPTCHA Interaction as Extended POMDP}

Solving a CAPTCHA as a GUI agent is an interactive process: the solver repeatedly observes a live webpage under partial information, maintains an internal task state, performs reasoning to make decision, and executes low-level browser actions. Our design goal is to exploit a persistent human–agent gap in this observation–memory–decision–action loop: humans can often infer the correct latent task state and act reliably with effective deliberation, whereas modern GUI agents frequently fail due to brittleness in perception, state maintenance, and perception-to-action alignment under realistic interaction constraints. We use an extended POMDP formulation to make this loop explicit and to organize the failure modes that Next-Gen CAPTCHAs intentionally amplify.

We model the GUI-agent as an extended POMDP:
\begin{equation}
\mathcal{W}
= (S,O,X,A_{\text{web}},A_{\text{think}},Z,
T_{\text{env}},U,R,\kappa),
\end{equation}

At step $t$, the live webpage has state $s_t\in S$, the agent receives an observation
$o_t\sim Z(\cdot\mid s_t)$, maintains an internal workspace $x_t\in X$, and selects
$a_t=(a_t^{\text{web}},a_t^{\text{think}})\in A_{\text{web}}\times A_{\text{think}}$,
where $a_t^{\text{web}}$ are browser actions (click/scroll/type/drag) and
$a_t^{\text{think}}$ is internal deliberation (e.g., a backbone thinking mode).
The webpage evolves as $s_{t+1}\sim T_{\text{env}}(\cdot\mid s_t,a_t^{\text{web}})$, and the
agent updates its workspace by
\begin{equation}
x_{t+1}=U(x_t,o_t,a_t^{\text{think}}),
\end{equation}
We view $U$ as comprising an observation-to-memory write and an optional deliberation step; when no
extra deliberation is invoked ($a_t^{\text{think}}=\varnothing$), $U$ reduces to a memory-only update.

Each CAPTCHA instance induces an initial state distribution $\rho$ over $S$ and a short horizon $H$
(termination when the verifier returns pass/fail). The verifier provides a terminal reward
$R\in\{0,1\}$ and we measure deliberation cost $c_t=\kappa(a_t^{\text{think}})$ (tokens/api cost).

\noindent{\bf Instantiation in our main experiments.}
Although $\mathcal{W}$ is a general formulation applicable to multiple GUI-agent stacks, our main results use browser-use as the reference
integration and restrict the observation payload to what is actually exposed at test time.
Concretely, we use
\begin{equation}
o_t=(I_t, D_t, \mathrm{meta}_t),
\end{equation}
where $I_t$ is a screenshot, $\mathrm{meta}_t$ includes URL and viewport/page statistics, and
$D_t$ is a filtered, DOM-derived interaction view (indexed interactive elements and puzzle-specific fields such as
window.currentPuzzle), not the full DOM tree.
We do not provide additional overlay/SoM annotation channels in the observation.

\noindent{\bf Human--agent cognitive decision gaps.}
Our goal is the standard CAPTCHA goal: easy for humans, hard for GUI agents. 
We start from general, recurring human--agent gaps reported in prior works and observations~\citep{upadhyay2025time}, and we turn them into concrete CAPTCHA families.
In the POMDP view, this means designing instances where success depends on reliably extracting task-relevant information from observations $o_t$, maintaining the correct task state $x_t$ across steps, and executing the appropriate browser action $a_t^{\text{web}}$.
We then instantiate these general gaps under our evaluation setting $\mathcal{W}$ (i.e., the observation payload and action interface exposed at test time), which yields the targeted gap categories below.

\noindent\textbf{(G1) Scene-structure inference (observation interpretation and grounding).}
We target failure modes where an agent cannot reliably extract the task-relevant structure of the page from its observations under partial observability, e.g., mis-parsing layered depth cues, reflections, or shadow geometry, or grounding the instruction to the wrong visual entity~\citep{pothiraj2025capture,liu2025beyond,lee2025perspective,motamed2025generative}.
In $\mathcal{W}$, this corresponds to brittleness of \emph{observation interpretation} under the observation channel $Z(\cdot\mid s_t)$: even when the underlying page state is fixed, small ambiguities or nuisance factors in $o_t$ lead to inconsistent task-relevant features, yielding an incorrect internal decision state $x_t$ and downstream actions.

\noindent\textbf{(G2) Temporal integration (multi-step evidence accumulation).}
We include tasks where the decisive information is not contained in any single snapshot, but is revealed only through interaction over multiple steps (e.g., motion cues or sequential reveals)~\citep{upadhyay2025time,bordes2025intphys,yuan2025videorefer}. In $\mathcal{W}$, success requires using the workspace update $x_{t+1}=U(x_t,o_t,a_t^{\text{think}})$ to
integrate evidence across steps, rather than reacting myopically to a single $o_t$. These tasks therefore stress whether the agent can stably integrate partial evidence across time into $x_t$.

\noindent\textbf{(G3) Numerosity and discrete invariants (decision-boundary sensitivity).}
We include tasks whose correct answer depends on discrete quantities or invariants (counts, parity, path endpoints), so that small perceptual errors can flip the final decision~\cite{guo2025your,weng2025visnumbench,Tamarapalli2025CountQAHW}.
In $\mathcal{W}$, this manifests as high sensitivity of the verifier reward $R$ to discrete task variables $g(s_t)$: small misreadings of $o_t$ can change the inferred value of $g(s_t)$ and thereby switch the rewarded outcome.

\noindent\textbf{(G4) Latent-state tracking (working-memory consistency).}
We include tasks that require carrying intermediate variables across steps, such as partial counts, orientations, or rule states, that may not be fully re-observable later~\citep{zhang2024working,huang2025llms}. 
In $\mathcal{W}$, this corresponds to maintaining these intermediates within $x_t$ and updating them consistently via the memory dynamics,
\begin{equation}
x_{t+1}=U(x_t,o_t,\varnothing).
\end{equation}
so that subsequent decisions and interactions remain coherent over the episode.

\noindent\textbf{(G5) Perception-to-action alignment (robust low-level execution).}
Distinct from perceptual understanding, we stress whether an agent can reliably translate a correct internal decision into the correct browser interaction~\citep{cheng2024seeclick,li2025screenspot}.
In $\mathcal{W}$, this is sensitivity in selecting and executing browser actions $a_t^{\text{web}}\in A_{\text{web}}$ such that intended transition is realized under the environment dynamics $T_{\text{env}}$ (e.g., precise drag targets), while remaining within practical deliberation cost $\kappa$.

\subsection{Data Curation Pipeline}
\label{sec:data_curation}

\begin{figure}[t]
    \centering
    \includegraphics[width=0.8\linewidth]{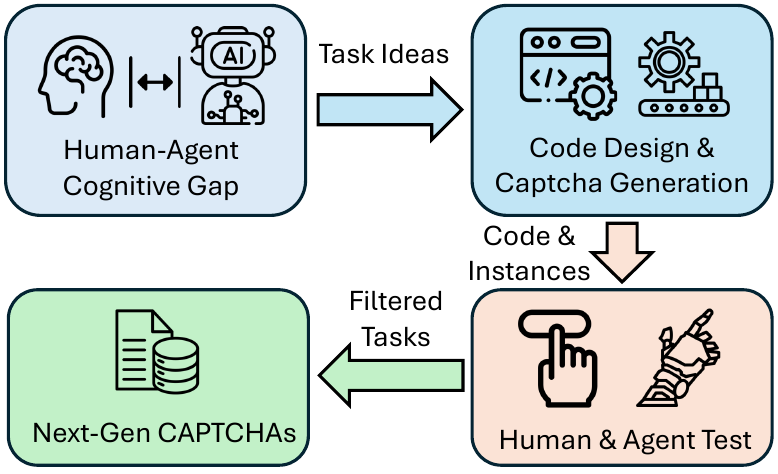}
    \vspace{-0.6em}
    \caption{
    \textbf{Next-Gen CAPTCHA data curation pipeline.}
    CAPTCHA task families are first designed to target human--agent cognitive gaps,
    then instantiated via procedural generation with rule-based verification.
    A lightweight model-based stress test and human test filters families that remain difficult
    for GUI agents while being human-solvable, yielding a curated benchmark
    sampled from a continuously generative defense system.
    }
    \label{fig:data_curation_pipeline}
    \vspace{-0.6em}
\end{figure}

We build the benchmark as a curated snapshot of a larger CAPTCHA defense system.
Starting from the human--agent gaps described above, we formed an initial pool of candidate CAPTCHA task families, informed by both classic CAPTCHA design and recent studies~\citep{upadhyay2025time}. For each candidate task family, we implemented an end-to-end task: a web interface, an instance generator when applicable, where the puzzle rules and solution are encoded directly in the generation process, ensuring correctness by construction rather than by an external checker.

Before running any model-based filtering, we conducted a preliminary manual verification to confirm that the rule-based generation behaves as intended and that the generated puzzles are interpretable and human-friendly and the task is genuinely usable as a CAPTCHA: instructions are unambiguous, interactions work as intended, and typical users can solve without domain knowledge. We then produced a small pilot set for each task family: 20 generated instances for generative families, and comparable curated examples for the few non-generated ones.

To decide which families are worth keeping, we used a lightweight but strong model (Gemini 3 Flash) as a first-pass stress test. We retained a family only when it remained difficult for the model (Pass@1 below 30\% on the pilot set) while being easy for humans (above 90\% success on a randomly sampled subset of 10 pilot instances). This filtering yielded 27 vision--language CAPTCHA families. Among them, 25 families are fully script-generated with automatic checking, while 2 utilize curated human-crafted instances.
From the retained families, we construct a main test set of 519 puzzles for evaluation. We also provide a budget-friendly subset by sampling 5 puzzles per family (135 total) to support quick comparisons under limited query budgets, while preserving coverage across task types.

\noindent\textbf{Summary of Next-Gen CAPTCHAs design objectives and properties.} Table~\ref{nextgen_summary} summarizes the key objectives and principles behind our Next-Gen CAPTCHAs: they are generative with rule-defined ground truth, cover diverse interaction bottlenecks, and remain human-friendly.
These properties shift CAPTCHAs away from static, decomposable logic and toward tasks that require robust interactive grounding and manipulation.

\begin{table}[h]
\centering
\setlength{\tabcolsep}{2.5pt}
\renewcommand{\arraystretch}{1.05}
\caption{\textbf{Next-Gen CAPTCHA design summary (aligned with Cognitive-Gap categories).}}

{\small
\resizebox{1.\linewidth}{!}{
\begin{tabular}{
  >{\raggedright\arraybackslash}p{0.29\linewidth}
  >{\raggedright\arraybackslash}p{0.55\linewidth}
  @{\hspace{2pt}}
  *{5}{>{\centering\arraybackslash}p{0.02\linewidth}}
}
\toprule
\multirow{2}{*}{\makecell[c]{\textbf{Key property}}} &
\multirow{2}{*}{\makecell[c]{\textbf{Summary}}} &
\multicolumn{5}{c}{\textbf{Gap}} \\
\cmidrule(lr){3-7}
& & {\scriptsize G1} & {\scriptsize G2} & {\scriptsize G3} & {\scriptsize G4} & {\scriptsize G5} \\
\midrule

Generative &
Procedural generation; scalable; resists memorization. &
{\scriptsize $\cdot$} & {\scriptsize $\cdot$} & {\scriptsize $\cdot$} & {\scriptsize $\cdot$} & {\scriptsize $\cdot$}
\\

Rule-defined GT &
GT verified by generation rules; no manual labels. &
{\scriptsize $\cdot$} & {\scriptsize $\cdot$} & {\scriptsize $\cdot$} & {\scriptsize $\cdot$} & {\scriptsize $\cdot$}
\\

Diverse coverage &
Covers G1--G5 families; avoids single-strategy solvers. &
{\scriptsize $\bullet$} & {\scriptsize $\bullet$} & {\scriptsize $\bullet$} & {\scriptsize $\bullet$} & {\scriptsize $\bullet$}
\\

Human-friendly &
High human success without domain knowledge. &
{\scriptsize $\cdot$} & {\scriptsize $\cdot$} & {\scriptsize $\cdot$} & {\scriptsize $\cdot$} & {\scriptsize $\cdot$}
\\

Agent-defensive &
Stresses realistic GUI interaction; mainly G5$^{+}$. &
{\scriptsize $\circ$} & {\scriptsize $\cdot$} & {\scriptsize $\cdot$} & {\scriptsize $\circ$} & {\scriptsize $\bullet$}
\\

Cognitive-Gap targeting &
Interactive bottlenecks to maximize human--agent separation. &
{\scriptsize $\bullet$} & {\scriptsize $\bullet$} & {\scriptsize $\bullet$} & {\scriptsize $\bullet$} & {\scriptsize $\bullet$}
\\

\bottomrule
\end{tabular}
}
\vspace{2pt}
\begin{minipage}{\linewidth}
\footnotesize
\textit{Legend:} $\bullet$ primary;\ $\circ$ secondary;\ $\cdot$ not targeted.\\
\textit{G1--G5 exemplars:} \textbf{G1} Mirror;\ \textbf{G2} Structure From Motion;\ \textbf{G3} Hole Counting;\ \textbf{G4} Box Folding;\ \textbf{G5} Static Jigsaw.\ 
\textit{Full family catalog:} see Table~\ref{tab:captcha_families}.\ 
$^{+}$ also stresses secondary gaps (e.g., G1/G4).
\end{minipage}
}
\label{nextgen_summary}
\vspace{-1.2em}
\end{table}

\begin{table*}[t]
\centering
\caption{Performance of different MLLM backbones within the Browser-Use baseline agent on Next-Gen Captcha.
Darker ``\ColorSquare{VintageBlueF}'' indicates higher success rate@1 and darker ``\ColorSquare{CreamF}'' indicates higher cost.}
\label{main_table}

\scriptsize
\setlength{\tabcolsep}{3pt}
\renewcommand{\arraystretch}{1.0}
\resizebox{.93\linewidth}{!}{
\begin{tabular}{@{}l c c c c c c c@{}}
\toprule
\textbf{Captcha Type}
& \textbf{Human}
& \makecell{\textbf{GPT-5.2}\\\textbf{xHigh}}
& \makecell{\textbf{Gemini-3}\\\textbf{Flash-High}}
& \makecell{\textbf{Claude-Opus4.5}\\\textbf{Extend-ThinkingHigh}}
& \makecell{\textbf{Gemini-3}\\\textbf{Pro-High}}
& \makecell{\textbf{Doubao-Seed-1.8}\\\textbf{Thinking-HighEffort}}
& \makecell{\textbf{Qwen3-VL-Plus}\\\textbf{ThinkingHigh}} \\
\midrule

\textbf{Avg Pass@1 (\%)}
& \textbf{98.8}
& \cellcolor{VintageBlueG}\textcolor{white}{\textbf{5.9}}
& \cellcolor{VintageBlueF}\textcolor{white}{\textbf{3.2}}
& \cellcolor{VintageBlueE}\textcolor{white}{\textbf{3.0}}
& \cellcolor{VintageBlueD}\textcolor{white}{\textbf{1.3}}
& \cellcolor{VintageBlueC}\textcolor{white}{\textbf{1.3}}
& \cellcolor{VintageBlueB}\textcolor{white}{\textbf{0.9}} \\
\textbf{Total Cost (\$)}
& --
& \cellcolor{CreamG}\textcolor{white}{\textbf{3122.3}}
& \cellcolor{CreamB}\textcolor{white}{\textbf{6.5}}
& \cellcolor{CreamF}\textcolor{white}{\textbf{224.0}}
& \cellcolor{CreamE}\textcolor{white}{\textbf{50.1}}
& \cellcolor{CreamC}\textcolor{white}{\textbf{26.6}}
& \cellcolor{CreamD}\textcolor{white}{\textbf{28.3}} \\

\midrule
3D\_Viewpoint                 & 100.0 & 0.0 & 0.0 & 0.0 & 0.0 & 0.0 & 0.0 \\
Backmost\_Layer               & 100.0 & \cellcolor{VintageBlueAA}20.0 & 0.0 & 0.0 & 0.0 & 0.0 & 0.0 \\
Box\_Folding                  & 100.0 & \cellcolor{VintageBlueAA}20.0 & 0.0 & 0.0 & 0.0 & 0.0 & 0.0 \\
Color\_Counting               & 100.0 & \cellcolor{VintageBlueAA}40.0 & \cellcolor{VintageBlueAA}5.0 & 0.0 & 0.0 & 0.0 & 5.0 \\
Dice\_Roll\_Path              & 100.0 & 0.0 & \cellcolor{VintageBlueAA}15.0 & 0.0 & 0.0 &\cellcolor{VintageBlueAA} 5.0 & \cellcolor{VintageBlueAA}15.0 \\
Dynamic\_Jigsaw               & 100.0 & 0.0 & 0.0 & 0.0 & 0.0 & 0.0 & 0.0 \\
Hole\_Counting                & 100.0 & 0.0 & 0.0 & 0.0 & 0.0 & 0.0 & 0.0 \\
Structure\_From\_Motion       & 90.0 & 0.0 & 0.0 & 0.0 & \cellcolor{VintageBlueAA}5.0 & 0.0 & 0.0 \\
Subway\_Paths                 & 100.0 & 0.0 & 0.0 & 0.0 & 0.0 & 0.0 & 0.0 \\
Temporal\_Object\_Continuity  & 100.0 & 0.0 & 0.0 & 0.0 & 0.0 & 0.0 & 0.0 \\
Trajectory\_Recovery          & 100.0 & 0.0 & 0.0 & 0.0 & 0.0 & 0.0 & 0.0 \\
Illusory\_Ribbons             & 100.0 & 0.0 & 0.0 & 0.0 & 0.0 & 0.0 & 0.0 \\
Layered\_Stack                & 100.0 & 0.0 & 10.0 & 0.0 & 0.0 & 0.0 & 0.0 \\
Mirror                        & 90.0 & \cellcolor{VintageBlueAA}20.0 & \cellcolor{VintageBlueAA}18.2 & 0.0 & \cellcolor{VintageBlueAA}9.1 & \cellcolor{VintageBlueAA}9.1 & 0.0 \\
Multi\_Script                  & 100.0 & 0.0 & 0.0 & 0.0 & 0.0 & 0.0 & 0.0 \\
Occluded\_Pattern\_Counting   & 100.0 & \cellcolor{VintageBlueAA}20.0 & 0.0 & 0.0 & \cellcolor{VintageBlueAA}5.0 & \cellcolor{VintageBlueAA}15.0 & 0.0 \\
Red\_Dot                      & 100.0 & 0.0 & \cellcolor{VintageBlueAA}15.0 & \cellcolor{VintageBlueAA}20.0 & 0.0 & 0.0 & 0.0 \\
Rotation\_Match               & 100.0 & \cellcolor{VintageBlueAA}20.0 & 0.0 & 0.0 & 0.0 & 0.0 & 0.0 \\
Shadow\_Direction             & 100.0 & \cellcolor{VintageBlueAA}20.0 & 0.0 & 0.0 & 0.0 & 0.0 & 0.0 \\
Shadow\_Plausible             & 100.0 & 0.0 & \cellcolor{VintageBlueAA}12.5 & 0.0 & 0.0 & 0.0 & 0.0 \\
Spooky\_Circle\_Grid          & 100.0 & 0.0 & 0.0 & 0.0 & \cellcolor{VintageBlueAA}5.0 & 0.0 & 0.0 \\
Spooky\_Jigsaw                & 90.0 & 0.0 & 0.0 & 0.0 & 0.0 & 0.0 & 0.0 \\
Spooky\_Shape\_Grid           & 100.0 & 0.0 & 0.0 & 0.0 & 0.0 & 0.0 & 0.0 \\
Spooky\_Size                  & 100.0 & 0.0 & \cellcolor{VintageBlueAA}5.0 & 0.0 & \cellcolor{VintageBlueAA}10.0 & 0.0 & \cellcolor{VintageBlueAA}5.0 \\
Spooky\_Text                  & 100.0 & 0.0 & 0.0 & 0.0 & 0.0 & 0.0 & 0.0 \\
Static\_Jigsaw                & 100.0 & 0.0 & 0.0 & \cellcolor{VintageBlueAA}60.0 & 0.0 & 0.0 & 0.0 \\
Spooky\_Circle                & 100.0 & 0.0 & \cellcolor{VintageBlueAA}5.0 & 0.0 & 0.0 & \cellcolor{VintageBlueAA}5.0 & 0.0 \\
\bottomrule
\end{tabular}
}
\end{table*}

\section{Experiments and Analysis}
\subsection{Experimental Settings}
We report Pass@1 under a fixed interactive evaluation protocol on live web CAPTCHAs.
Unless stated otherwise, all main results use the \texttt{Browser-Use} GUI-agent as the default framework, running each CAPTCHA as a real-browser episode with state reset between puzzles.
At each step, the agent observes a multimodal page state (viewport screenshot + DOM-derived interactive elements + lightweight metadata) and outputs structured browser actions (e.g., click, type, drag, scroll) executed by Playwright in visible mode; success is determined by the platform’s automatic verification signals. More details are in Appendix.~\ref{appendix:experiment_details}.

\subsection{Results on Next-Gen's Testing Benchmark}

Table~\ref{main_table} summarizes the performance of different MLLM backbones under the same Browser-Use baseline agent on Next-Gen's testing benchmark. Humans achieve near-ceiling performance (98.8\% Pass@1), while all evaluated MLLM-based agents are successfully defended by our CAPTCHAs, with the best-performing backbone reaching only 5.9\% Pass@1. This stark discrepancy indicates that Next-Gen CAPTCHAs preserve a substantial security margin against current GUI-enabled agents. We also report the cost of each MLLMs backbone on Next-Gen CAPTCHAs, with more detailed analysis in Section~\ref{cost_section}.

Among agent backbones, GPT-5.2-xHigh yields the highest Pass@1 (5.9\%), followed by Gemini-3-Flash-High (3.2\%) and Claude-Opus4.5-Extended-ThinkingHigh (3.0\%). Gemini-3-Pro-High (1.3\%), Doubao-Seed-1.8-Thinking-HighEffort (1.3\%), and Qwen3-VL-Plus-ThinkingHigh (0.9\%) cluster around \(\sim\)1\%. Notably, these are widely regarded as the strongest reasoning MLLMs up-to-date, yet their success remains low, suggesting  dominant errors arise from interactive deployment bottlenecks rather than perception or language reasoning alone: precise spatial grounding, maintaining intermediate state across steps, and robust action execution within the UI.
Consistent with Fig.~\ref{correlation_analysis}, agents often attend to the correct region but choose wrong \emph{action primitive} (e.g., clicking \texttt{Submit} instead of dragging/rotating or taking more screenshots for further analysis), so additional steps and budget do not increase success. From a security perspective, residual Pass@1 can be further reduced by the rate limits and interaction friction, while attackers' pay attempts and long trajectories create a favorable economic asymmetry for the defenders.

\subsection{Ablation and Analysis}

\paragraph{Cost-Efficiency and Economic Asymmetry.}
\label{cost_section}
\begin{figure}[!h]
\centering
\vspace{-0.1in}
\includegraphics[width=1.0\linewidth]{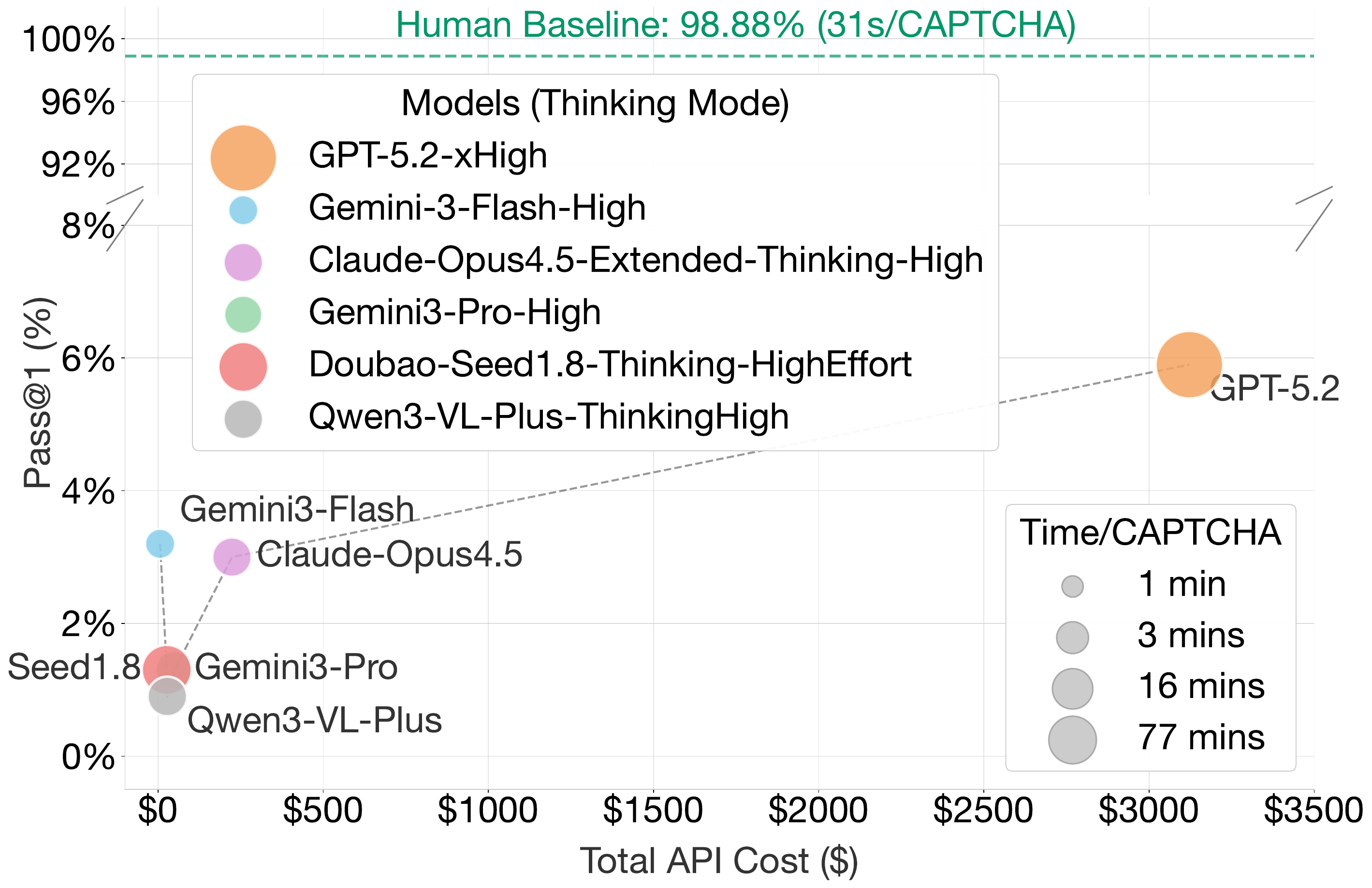}  
\vspace{-1.5em}  
\caption{\textbf{Cost--accuracy--latency trade-off on Next-Gen.}
We plot each model's Pass@1 (y-axis) against total API cost (x-axis); bubble size indicates average time per CAPTCHA.
}
\vspace{-.5em}  
\label{cost_effective}
\end{figure}
We evaluate the economic viability of attacking Next-Gen CAPTCHAs by analyzing the trade-off between attack success rates (Pass@1), API costs (billed usage) and time to finish the evaluation in a live API setting across varying MLLMs.

\begin{table}[t]
\centering
\small
\setlength{\tabcolsep}{10pt}
\renewcommand{\arraystretch}{0.95}
\caption{\textbf{Ablation on agent frameworks (overall).}
Overall Pass@1 (\%) on the 135-puzzle subset of the Next-Gen CAPTCHA benchmark under different GUI-agent frameworks. Backbone MLLM and protocol are held constant (Claude-Opus4.5).}
\begin{tabular}{l c}
\toprule
\textbf{Framework} & \textbf{Overall Pass@1 (\%)} \\
\midrule
CrewAI        & 0.00 \\
Browser-Use   & 1.48 \\
Claude Cowork & 4.44 \\
\bottomrule
\end{tabular}
\vspace{-0.6em}
\label{tab:framework_ablation}
\end{table}

As illustrated in Fig.~\ref{cost_effective}, we observe a distinct lack of correlation between computational expenditure and task success, effectively breaking the scaling laws typically seen in static benchmarks. While the human baseline demonstrates a near-perfect success rate of 98.8\% with an average completion time of just 31 seconds, even the most advanced frontier models fail to achieve meaningful penetration. Notably, GPT-5.2-Xhigh, utilizing its maximum reasoning capacity, incurs an exorbitant cost around \$3,122 to process the evaluation set yet achieves a Pass@1 rate of only 5.9\%. This represents a highly unfavorable cost–benefit for potential attackers, as scaling inference costs by orders of magnitude yields negligible gains in solvability.

\begin{figure}[h]
    \centering
    \vspace{-0.1em}
    \includegraphics[width=0.93\linewidth]{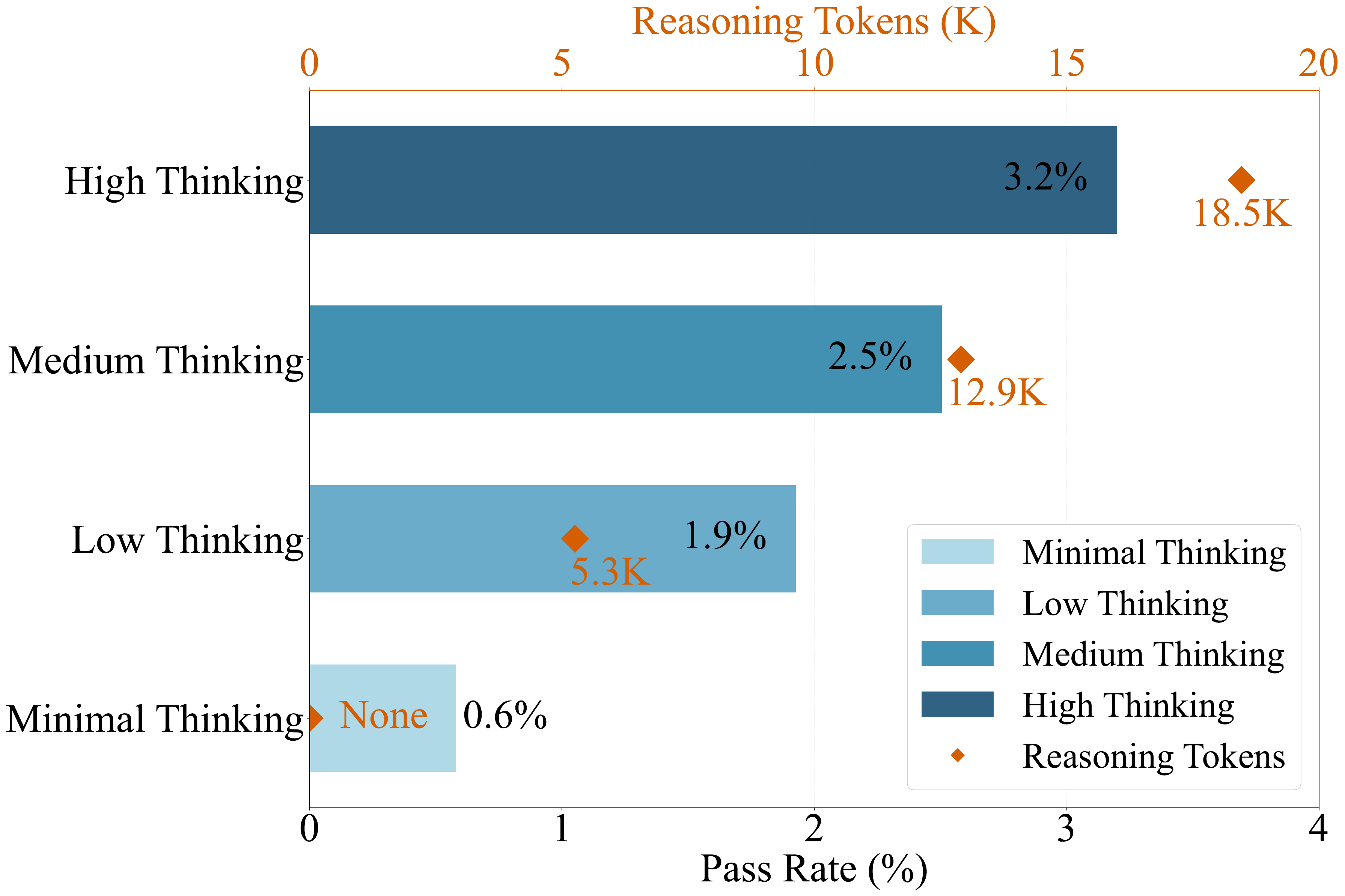}
    \vspace{-0.6em}
    \caption{\textbf{Thinking-mode ablation on Gemini-3-Flash.}
    \textbf{Left:} Pass@1 under four thinking modes (Minimal/Low/Medium/High).
    \textbf{Right:} the corresponding average reasoning-token usage.
    While higher reasoning budget yields consistent gains, improvements quickly saturate and remain far from reliable success, indicating that Next-Gen CAPTCHAs are not reasoning-heavy and that interactive perception--action bottlenecks dominate failures.}
    \label{fig:thinking_modes}
    \vspace{-.28em}
\end{figure}

Furthermore, the latency overhead introduced by agentic reasoning serves as a secondary defense layer. The bubble sizes in Fig.~\ref{cost_effective} indicate that high-reasoning models can require anywhere from 16 to 77 minutes to attempt a single complex puzzle, a timeframe that renders real-time automated attacks operationally infeasible compared to the rapid sub-minute performance of human users. This confirms that Next-Gen CAPTCHAs successfully impose a severe economic and temporal asymmetry, ensuring that automated attacks remain both technically intractable and financially ruinous regardless of the compute budget allocated.

\noindent\textbf{Reasoning Marginally Helps, but Next-Gen Is Not Reasoning-Heavy.}
We further analyze whether allocating more deliberate reasoning budget can meaningfully improve an agent's ability to solve Next-Gen CAPTCHAs.
As shown in Fig.~\ref{fig:thinking_modes}, increasing the thinking mode of Gemini3-flash consistently improves Pass@1, indicating that reasoning is indeed helpful.
However, the improvement is small and quickly saturates: Pass@1 rises only from 0.6\% (Minimal) to 1.9\% (Low), 2.5\% (Medium), and 3.2\% (High), despite a large increase in reasoning-token usage (from none to 5.3K/12.9K/18.5K tokens for Low/Medium/High).
This pattern suggests that Next-Gen CAPTCHAs are \emph{not} primarily limited by long-horizon textual reasoning; instead, failures are dominated by interactive bottlenecks such as fine-grained visual--spatial grounding, temporal evidence integration, short-term state maintenance across steps, and robust low-level action execution.
Consequently, our Next-Gen families remain strongly agent-defensive even against today's reasoning-augmented models: simply ``thinking more'' is insufficient to reliably pass.

\noindent\textbf{Agent Frameworks Matter, but Do Not Close the Gap.}
To test whether our results depend on the agent orchestration layer, we ablate the underlying GUI-agent frameworks on the 135-puzzle Next-Gen CAPTCHA subset, while holding the browser interface, evaluation protocol, and backbone MLLM fixed (Claude-Opus4.5). Table~\ref{tab:framework_ablation} reports the overall Pass@1 for three representative, up-to-date frameworks. We observe a clear separation: CrewAI achieves 0.00\% Pass@1, Browser-Use reaches 1.48\%, and Claude Cowork attains 4.44\%.
These differences indicate that the interaction stack and action execution quality can measurably affect outcomes. However, the absolute performance remains uniformly low across all frameworks, showing that improved orchestration alone does not eliminate the fundamental difficulty posed by Next-Gen CAPTCHAs. 

\vspace{-0.05in}
\section{Conclusion}
% \vspace{-0.02in}

 We present \textbf{Next-Gen CAPTCHAs}, a systematic defense framework designed to secure the web against the advanced autonomous GUI-enabled agents. Our investigation reveals that the arms race between CAPTCHAs and AI has reached a tipping point: recent advancements in reasoning-heavy MLLMs, such as GPT-5.2 and Gemini3-Pro, have effectively trivialized existing logic-based benchmarks, achieving near-perfect pass rates on previously secure tasks. This collapse of the bot-hard assumption exposes critical vulnerabilities in the current web infrastructure, where agents can now navigate, reason, and act with human-like proficiency. To counter this, we introduce a paradigm shift from static difficulty to dynamic cognitive divergence. By exploiting the ``Cognitive Gap'', the inherent asymmetry between human intuition and the over-segmented, costly reasoning processes of current agents, we successfully restore the distinction between biological and artificial users.
 Moreover, our pipeline enables continual refresh at scale and provides a practical path for deploying CAPTCHA services that can defend real-world web systems against advanced GUI agents.

\section*{Impact Statement}
This paper studies the security implications of modern multimodal GUI agents on existing CAPTCHA systems and proposes Next-Gen CAPTCHAs together with an evaluation platform to measure agent robustness in realistic web-interaction settings.
Our goal is to advance machine learning and its safe deployment by identifying concrete failure modes of current agents, and by providing a human-friendly, scalable defense mechanism against automated abuse.

\textbf{Potential benefits.}
The primary expected benefit is improved web security and trustworthiness: stronger agent-resistant CAPTCHAs can help mitigate automated fraud, credential stuffing, spam, mass account creation, scraping, and other forms of abuse.
Our benchmark and platform can also support research on safer and more reliable interactive agents by enabling reproducible evaluation of perception--action grounding, memory, and manipulation under real-world constraints.

\textbf{Potential risks and misuse.}
As with many security-related publications, our work could be dual-use.
Detailed descriptions of CAPTCHA design and solver behavior may aid adversaries in developing more effective bypass strategies, or in stress-testing deployed defenses.
To mitigate this risk, our focus is on principles for agent-defensive challenge design and on evaluation methodology rather than releasing attack tooling; moreover, procedural generation enables rapid instance diversification, which can reduce the value of static memorization-based attacks.

\textbf{Fairness, accessibility, and user burden.}
CAPTCHAs can impose friction and may disproportionately impact users with disabilities or limited access to audio/visual modalities.
We therefore emphasize human friendliness in design and include a human study to measure completion time and success rate; nevertheless, we acknowledge that accessibility and usability require ongoing attention for real deployments (e.g., alternative modalities, localization, and adaptive difficulty).
Overall, we believe the potential benefits to web security and the responsible evaluation of GUI agents outweigh the foreseeable risks, and we encourage future work on accessibility-aware and privacy-preserving deployment of Next-Gen CAPTCHA defenses.

\section*{Acknowledgements}
This work is supported by the MBZUAI-WIS Joint Program for Artificial Intelligence Research.

\bibliography{reference}
\bibliographystyle{icml2026}  

%%%%%%%%%%%%%%%%%%%%%%%%%%%%%%%%%%%%%%%%%%%%%%%%%%%%%%%%%%%%%%%%%%%%%%%%%%%%%%%
%%%%%%%%%%%%%%%%%%%%%%%%%%%%%%%%%%%%%%%%%%%%%%%%%%%%%%%%%%%%%%%%%%%%%%%%%%%%%%%
% APPENDIX
%%%%%%%%%%%%%%%%%%%%%%%%%%%%%%%%%%%%%%%%%%%%%%%%%%%%%%%%%%%%%%%%%%%%%%%%%%%%%%%
%%%%%%%%%%%%%%%%%%%%%%%%%%%%%%%%%%%%%%%%%%%%%%%%%%%%%%%%%%%%%%%%%%%%%%%%%%%%%%%
\newpage
\appendix
\onecolumn

\section*{\Large{Appendix}}

%%%%%%%%%%%%%%%%%%%%%%%%%%%%%%%%%%%%%%%%%%%%%%%%%%%%%%%%%%%%%%%%%%%%%%%%%%%%%%%
%%%%%%%%%%%%%%%%%%%%%%%%%%%%%%%%%%%%%%%%%%%%%%%%%%%%%%%%%%%%%%%%%%%%%%%%%%%%%%%
\section{More Details of Experimental Settings}
\label{appendix:experiment_details}

We report Pass@1 under the evaluation protocol. Due to API budget constraints, we use a two-tier evaluation: we run Gemini3-Pro-High, Gemini3-Flash-High~\citep{gemini3}, Qwen3-VL-Plus-ThinkingHigh~\citep{bai2025qwen3}, and Doubao-Seed-1.8-Thinking-HighEffort~\citep{bytedance_seed_seed18_modelcard} on the full testing set of 519 puzzles.
We evaluate GPT-5.2-xHigh~\citep{gpt5.2} and Claude-Opus4.5-Extended-ThinkingHigh~\citep{claude4.5} on a 135-puzzle subset because max-reasoning closed-API inference exhibits substantial latency variance and occasional severe queuing delays at the step level, which makes running the full 519-puzzle suite prohibitively time and cost intensive (e.g, GPT-5.2-xHigh sometimes spends up to 1 hour to run 1 step). Unless stated otherwise, these are the default settings for all main results.

For GUI Agent framework, we evaluate all models using \texttt{Browser-Use}~\citep{browseruse_github} as the default agent framework for all main results and non-framework ablations, running each CAPTCHA as an interactive episode in a real browser.
 At each step $t$, the agent observes the current page state via $O_t=(I_t,D_t,\mathrm{meta}_t)$, including a viewport screenshot, a structured DOM-derived view of interactive elements (indexed for interaction), and lightweight context such as the current URL and task information. The model then returns a JSON response that specifies the next browser actions, which are executed to advance the episode. We run the browser through Playwright~\citep{playwright} in visible (non-headless) mode and reset the agent state for every puzzle. Success or failure is recorded using the platform’s built-in outcome signals.

We additionally evaluate alternative orchestration frameworks including \texttt{CrewAI}~\citep{crewai_github} and \texttt{Claude Cowork}~\citep{claudeCowork} under the standard protocol.

\section{Human Usability Evaluations}
\label{appendix:human_usability}

We report human usability results from two participant groups.

The Human column in Table~\ref{main_table} reports results from the CS-background human evaluation.
This evaluation uses 10 puzzle instances for most CAPTCHA families and 8 instances for \texttt{Shadow\_Plausible}, giving 268 human trials in total.
Participants were regular computer users.
No explicit time limit was imposed, and participants submitted once they were confident.
This evaluation achieves 98.8\% Pass@1. The average completion time is 31.1 seconds per puzzle.
The higher-time families are \texttt{3D\_Viewpoint} (120.8 seconds), \texttt{Spooky\_Jigsaw} (71.8 seconds), \texttt{Subway\_Paths} (71.4 seconds), and \texttt{Multi\_Script} (63.1 seconds); the remaining task families have average completion times below 60 seconds.

The second evaluation covers participants without CS backgrounds.
It includes 8 participants, each solving one instance from each of the 27 families, yielding 216 trials in total.
This evaluation achieves 85.2\% Pass@1.
Table~\ref{tab:noncs_human_profile} summarizes the participant attributes and per-participant Pass@1.
Together, the two evaluations report human performance for both CS/AI-background and non-CS/AI-background participant groups.

\begin{table}[h]
\centering
\small
\setlength{\tabcolsep}{4pt}
\renewcommand{\arraystretch}{1.05}
\resizebox{\linewidth}{!}{%
\begin{tabular}{l c c c c c c c c c}
\toprule
\textbf{Attribute} & \textbf{P1} & \textbf{P2} & \textbf{P3} & \textbf{P4} & \textbf{P5} & \textbf{P6} & \textbf{P7} & \textbf{P8} & \textbf{Overall} \\
\midrule
Education & Master & Bachelor & Primary School & Bachelor & Master & PhD & Bachelor & Primary School & -- \\
Age group & 20--30 & 20--30 & 10--20 & 80--90 & 50--60 & 50--60 & 20--30 & 10--20 & -- \\
Expertise & Finance & Psychology & N/A & Business & Literature & Finance & Media & N/A & -- \\
Pass@1 (\%) & 81.5 & 92.6 & 88.9 & 66.7 & 92.6 & 85.2 & 85.2 & 88.9 & 85.2 \\
\bottomrule
\end{tabular}%
}
\caption{\textbf{Non-CS/AI-background human usability evaluation.}
Each participant attempted one instance from each of the 27 CAPTCHA families; Overall reports aggregate Pass@1 across 216 trials.}
\label{tab:noncs_human_profile}
\end{table}

\section{Robustness to Family-Aware Hint Adaptation}
\label{appendix:adaptive_hint}

We evaluate a family-aware hint adaptation setting on the full 519-puzzle test set using Gemini-3-Flash-High with \texttt{Browser-Use}.
Each CAPTCHA family is paired with one solving strategy hint generated by Gemini-3-Pro-High from the corresponding generator logic, frontend interaction code, server-side verification logic, and browser-agent task description.
The hint describes observable webpage cues and potentially useful interactions for the family.

Under this family-aware hint adaptation attack, overall Pass@1 increases from 3.2\% to 6.4\%.
The gain is concentrated in a small number of families, especially \texttt{Red\_Dot}, while most CAPTCHA families remain at or near their baseline accuracy.
Overall, the Next-Gen CAPTCHA families remain robust under this hint-based adaptation setting.

\begin{table}[h]
\centering
\small
\setlength{\tabcolsep}{5pt}
\renewcommand{\arraystretch}{1.05}
\begin{tabular}{l c c c}
\toprule
\textbf{Puzzle type} & \textbf{Baseline} & \textbf{Hint} & \textbf{$\Delta$ pp} \\
\midrule
\textbf{Overall} & \textbf{3.2} & \textbf{6.4} & \textbf{+3.2} \\
\texttt{3D\_Viewpoint} & 0.0 & 0.0 & 0.0 \\
\texttt{Backmost\_Layer} & 5.0 & 0.0 & -5.0 \\
\texttt{Box\_Folding} & 10.0 & 5.0 & -5.0 \\
\texttt{Color\_Counting} & 0.0 & 5.0 & +5.0 \\
\texttt{Dice\_Roll\_Path} & 15.0 & 20.0 & +5.0 \\
\texttt{Dynamic\_Jigsaw} & 0.0 & 0.0 & 0.0 \\
\texttt{Hole\_Counting} & 0.0 & 5.0 & +5.0 \\
\texttt{Illusory\_Ribbons} & 0.0 & 0.0 & 0.0 \\
\texttt{Layered\_Stack} & 0.0 & 0.0 & 0.0 \\
\texttt{Mirror} & 9.0 & 0.0 & -9.0 \\
\texttt{Multi\_Script} & 5.0 & 0.0 & -5.0 \\
\texttt{Occluded\_Pattern\_Counting} & 0.0 & 5.0 & +5.0 \\
\texttt{Red\_Dot} & 15.0 & 100.0 & +85.0 \\
\texttt{Rotation\_Match} & 0.0 & 0.0 & 0.0 \\
\texttt{Shadow\_Direction} & 0.0 & 0.0 & 0.0 \\
\texttt{Shadow\_Plausible} & 0.0 & 12.5 & +12.5 \\
\texttt{Spooky\_Circle} & 20.0 & 5.0 & -15.0 \\
\texttt{Spooky\_Circle\_Grid} & 0.0 & 0.0 & 0.0 \\
\texttt{Spooky\_Jigsaw} & 0.0 & 0.0 & 0.0 \\
\texttt{Spooky\_Shape\_Grid} & 5.0 & 0.0 & -5.0 \\
\texttt{Spooky\_Size} & 5.0 & 15.0 & +10.0 \\
\texttt{Spooky\_Text} & 0.0 & 0.0 & 0.0 \\
\texttt{Static\_Jigsaw} & 0.0 & 0.0 & 0.0 \\
\texttt{Structure\_From\_Motion} & 0.0 & 0.0 & 0.0 \\
\texttt{Subway\_Paths} & 0.0 & 0.0 & 0.0 \\
\texttt{Temporal\_Object\_Continuity} & 0.0 & 0.0 & 0.0 \\
\texttt{Trajectory\_Recovery} & 0.0 & 0.0 & 0.0 \\
\bottomrule
\end{tabular}
\caption{\textbf{Robustness to family-aware hint adaptation.}
Pass@1 (\%) on the full 519-puzzle test set using Gemini-3-Flash-High with \texttt{Browser-Use}.}
\label{tab:adaptive_hint}
\end{table}

\section{Dynamic-Family Payload Measurements}
\label{appendix:network_requirements}

We measure the cold-load visual asset payload for CAPTCHA families that use GIF-based or other non-static visual assets.
These measurements characterize transferred asset size under a cold-load setting and provide deployment-relevant information for selecting task families across network and device conditions.
Table~\ref{tab:network_payload} summarizes the measured payloads.

\begin{table}[h]
\centering
\small
\setlength{\tabcolsep}{6pt}
\renewcommand{\arraystretch}{1.05}
\begin{tabular}{l c}
\toprule
\textbf{Task family} & \textbf{Avg. cold-load payload (MiB)} \\
\midrule
\texttt{Trajectory\_Recovery} & 0.47 \\
\texttt{Temporal\_Object\_Continuity} & 0.54 \\
\texttt{Dynamic\_Jigsaw} & 0.73 \\
\texttt{Spooky\_Circle\_Grid} & 3.40 \\
\texttt{Spooky\_Shape\_Grid} & 3.48 \\
\texttt{Spooky\_Text} & 3.67 \\
\texttt{Spooky\_Circle} & 3.86 \\
\texttt{Spooky\_Size} & 6.77 \\
\texttt{Structure\_From\_Motion} & 7.73 \\
\texttt{Spooky\_Jigsaw} & 9.23 \\
\bottomrule
\end{tabular}
\caption{\textbf{Cold-load payloads for GIF-based dynamic families.}
Payload is reported as the average amount of visual asset data loaded per puzzle instance under a cold-load setting.}
\label{tab:network_payload}
\end{table}

In deployment, the system can choose task families based on connection quality and device capability, cache or preload assets when appropriate, and serve static or lower-bandwidth families for settings where lighter assets are preferred.
These options allow deployments to balance task diversity, user experience, and system constraints while retaining the same rule-based verification interface.

\section{Details of Next-Gen CAPTCHAs}
\label{appendix:nextgen_details}
This appendix provides implementation and evaluation details for our Next-Gen CAPTCHA families. The design goal is to move beyond \emph{static, decomposable logic puzzles} and instead elicit failures that stem from the \emph{Cognitive Gap}: robust interactive grounding, manipulation planning, and correct execution of action primitives under realistic UI constraints.

\paragraph{Design principles.}

\begin{table*}[h]
\centering
\resizebox{\linewidth}{!}{
\begin{tabular}{@{}p{0.24\textwidth} p{0.58\textwidth} c p{0.14\textwidth} c c@{}}
\toprule
Name & Brief description & Generative & Answer type & Benchmark size (\#) & Targeted gaps \\
\midrule
Mirror & Find mirror options that do \emph{not} match the reflected reference & No & Select & 11 [5] & G1 \\
Dice\_Roll\_Path & Roll a die along a shown path; report the final top face & Yes & Numeric & 20 [5] & G3, G4, G5 \\
Shadow\_Plausible & Pick images with physically plausible shadows in a grid & No & Select & 8 [5] & G1 \\
Layered\_Stack & Select cells where top shape and counts in lower layers meet a rule & Yes & Select & 20 [5] & G1, G3 \\
Red\_Dot & Timed clicks on appearing red dots until hit quota & Yes & Click position & 20 [5] & G5 \\
Color\_Counting & Select the cells that meet the rule about the number of colours in the cell & Yes & Select & 20 [5] & G3 \\
Spooky\_Circle & Count circles only visible via motion-contrast noise & Yes & Numeric & 20 [5] & G2 \\
Spooky\_Size & Click largest/smallest target shape visible only via motion contrast & Yes & Click position & 20 [5] & G2, G5 \\
Box\_Folding & Choose the folded cube that matches the given net & Yes & Select & 20 [5] & G1, G4 \\
3D\_Viewpoint & Select all views showing the same colored-edge wireframe & Yes & Select & 20 [5] & G1, G4, G5 \\
Backmost\_Layer & Click cells where the backmost (occluded) shape matches reference & Yes & Select & 20 [5] & G1 \\
Dynamic\_Jigsaw & Drag/drop animated GIF pieces to complete a 3×3 jigsaw & Yes & Drag-and-drop & 20 [5] & G2, G4, G5 \\
Hole\_Counting & Count topological holes in presented glyphs/shapes & Yes & Numeric & 20 [5] & G1, G3 \\
Illusory\_Ribbons & Select cells containing exactly the target number of ribbon loops & Yes & Select & 20 [5] & G1, G3 \\
Occluded\_Pattern\_Counting & Count two specified shapes under a semi-transparent occluder & Yes & Numeric & 20 [5] & G1, G3 \\
Rotation\_Match & Select tiles of the most frequent shape, ignoring rotation/color/texture & Yes & Select & 20 [5] & G1, G4 \\
Shadow\_Direction & Match light-source direction from photorealistic shadows & Yes & Select & 20 [5] & G1 \\
Spooky\_Circle\_Grid & Count how many grid cells contain motion-contrast circles & Yes & Numeric & 20 [5] & G2, G3 \\
Spooky\_Jigsaw & Drag/drop motion-contrast pieces to complete the jigsaw & Yes & Drag-and-drop & 20 [5] & G2, G4, G5 \\
Spooky\_Shape\_Grid & Select spooky cells with the target shape and rotation direction & Yes & Select & 20 [5] & G2 \\
Spooky\_Text & Read and type text visible only via motion contrast & Yes & Text entry & 20 [5] & G2 \\
Static\_Jigsaw & Drag/drop static pieces to complete a jigsaw & Yes & Drag-and-drop & 20 [5] & G4, G5 \\
Structure\_From\_Motion & Select GIF cells whose dot motion reflects the same rigid 3D shape & Yes & Select & 20 [5] & G2 \\
Trajectory\_Recovery & Watch a reference trajectory GIF; select matching trajectory plots & Yes & Select & 20 [5] & G2, G4 \\
Multi\_Script & Select cells containing any target characters (multi-script, transformed) & Yes & Select & 20 [5] & G1 \\
Subway\_Paths & Select maps with the specified count of valid routes under stamp rules & Yes & Select & 20 [5] & G3, G4 \\
Temporal\_Object\_Continuity & Select GIF cells where identity changes behind occluders & Yes & Select & 20 [5] & G2, G4 \\
\midrule
Summary & Captcha family from our GUI-Agent Defense System (27 types) & Yes & Mixed & 519 [135] & G1--G5 \\
\bottomrule
\end{tabular}
}
\caption{\textbf{Overview of Next-Gen CAPTCHA families.}
Benchmark size is reported as \textbf{main [lite]} (lite: 5 instances per family).
“Generative” denotes programmatic instance generation by our code system; “Targeted gaps” refers to the primary gaps in Section~4.}
\label{tab:captcha_families}
\end{table*}

Across families, we follow a shared set of principles:
\begin{itemize}
    \item \textbf{Interactive grounding over static recognition.} Most tasks cannot be solved from a single screenshot plus short reasoning. Correct completion requires discovering actionable UI affordances and grounding targets throughout interaction.
    \item \textbf{Action-primitive dependence.} Success requires the correct primitive (e.g., drag-and-drop, long-press, multi-step selection, constrained ordering). Incorrect primitives lead to hard failure even with strong perception.
    \item \textbf{Non-local dependence.} Early actions constrain later feasibility (e.g., piece placement affects future options), reducing the effectiveness of myopic step-by-step workflows.
    \item \textbf{Feedback that is informative but non-leaking.} UI feedback supports legitimate users (e.g., highlighting valid drop zones) without collapsing the task into a trivial reward signal exploitable by brute-force agents.
    \item \textbf{Human-feasible time/effort.} Each instance is designed to be solvable by typical users quickly, while remaining brittle for agents that mis-ground actions or affordances.
\end{itemize}

\paragraph{Instance generation and diversity.}
Each family instantiates challenges by sampling from a parameterized generator:
\begin{itemize}
    \item \textbf{Content parameters} (visual themes, object sets, layouts) to prevent memorization and template matching.
    \item \textbf{Interaction parameters} (number of manipulable elements, allowed moves, constraints) to control difficulty.
    \item \textbf{Anti-shortcut constraints} that eliminate single-step solutions (e.g., enforcing multi-stage manipulation, requiring ordering constraints, or gating completion behind a correctly executed primitive).
\end{itemize}
We recommend maintaining a large instance pool with per-request randomization, and rotating asset sets periodically to reduce overfitting.

\paragraph{UI instrumentation and logging.}
For each attempt, we log a trajectory consisting of:
\begin{itemize}
    \item \textbf{Observations:} screenshots (or DOM render snapshots) at each step.
    \item \textbf{Actions:} primitive type (click/drag/scroll/keyboard), coordinates/targets, and timestamps.
    \item \textbf{Agent deliberation signals:} step count, duration, and any available token-related signals for model-based agents.
\end{itemize}
These logs support the correlation analyses in Fig.~\ref{correlation_analysis} and enable failure-mode categorization (e.g., wrong primitive, mis-grounded target, premature submission).

\paragraph{Success criteria and verification.}
Each Next-Gen family defines a crisp completion predicate that is verified server-side to prevent client-side spoofing. Examples include:
\begin{itemize}
    \item \textbf{State-based verification:} the final arrangement satisfies a constraint system (e.g., correct assembly/ordering).
    \item \textbf{Action-consistency checks:} required primitives occurred with valid targets (e.g., a drag sequence that ends in a valid drop zone).
    \item \textbf{Anti-replay:} per-instance nonces and short TTLs to prevent reuse of solutions.
\end{itemize}

\paragraph{Evaluation protocol.}
We evaluate Next-Gen families with browser-based agents under a consistent interaction budget:
\begin{itemize}
    \item \textbf{Budgets:} maximum steps and wall-clock duration per attempt.
    \item \textbf{Constraints:} no privileged access beyond what a normal user/automation script can observe (i.e., no direct DOM oracle for hidden labels).
    \item \textbf{Metrics:} Pass@1 as the primary metric, plus trajectory statistics (Steps, Duration, and reasoning-related signals when available) used in Fig.~\ref{correlation_analysis}.
\end{itemize}

\paragraph{Security and usability considerations.}
Next-Gen CAPTCHAs are intended to be:
\begin{itemize}
    \item \textbf{Agent-defensive:} brittle to incorrect primitives and shallow workflow decomposition.
    \item \textbf{User-friendly:} solvable with intuitive affordances and minimal cognitive load.
    \item \textbf{Robust in deployment:} server-side verification, rate limiting, and anomaly detection over trajectories to reduce large-scale automated attacks.
\end{itemize}

\subsection{Next-Gen CAPTCHA Examples Gallery}
\label{appendix:captcha_gallery}
The following gallery presents representative instances from our Next-Gen CAPTCHA families. Each example is intended to illustrate \emph{why} the family remains difficult for current GUI agents even when static perception is strong.

\begin{figure*}[h]
    \centering
    \includegraphics[width=1.\linewidth]{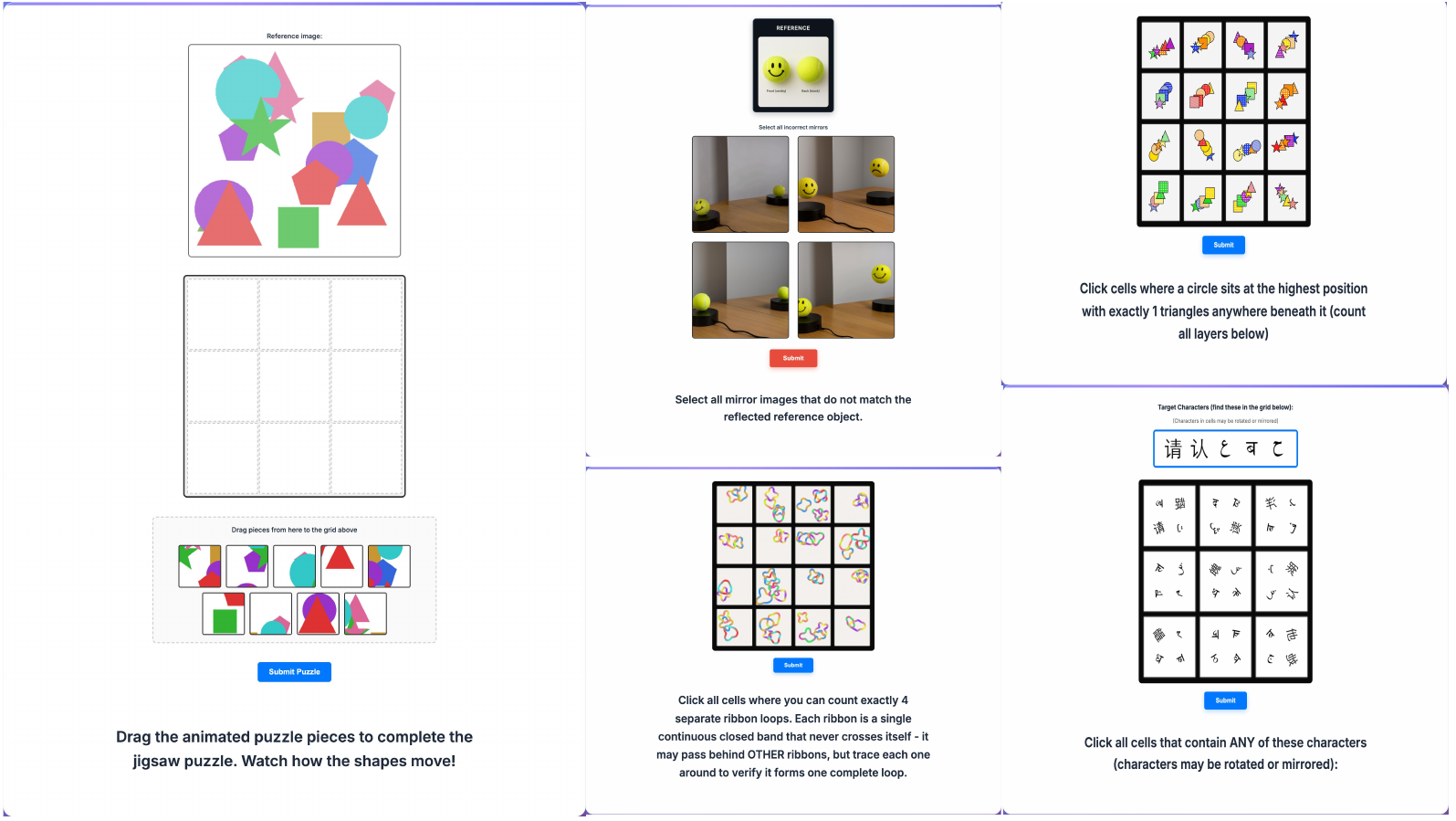}
    \vspace{-0.6em}
    \caption{\textbf{Five concrete examples of Next-Gen CAPTCHAs.} Dynamic Jigsaw, Mirror, Illusory Ribbons, Layered Stack and Multi Script.}
    \label{example1}
    \vspace{-.5em}
\end{figure*}

\begin{figure*}[t]
    \centering
    \includegraphics[width=1.\linewidth]{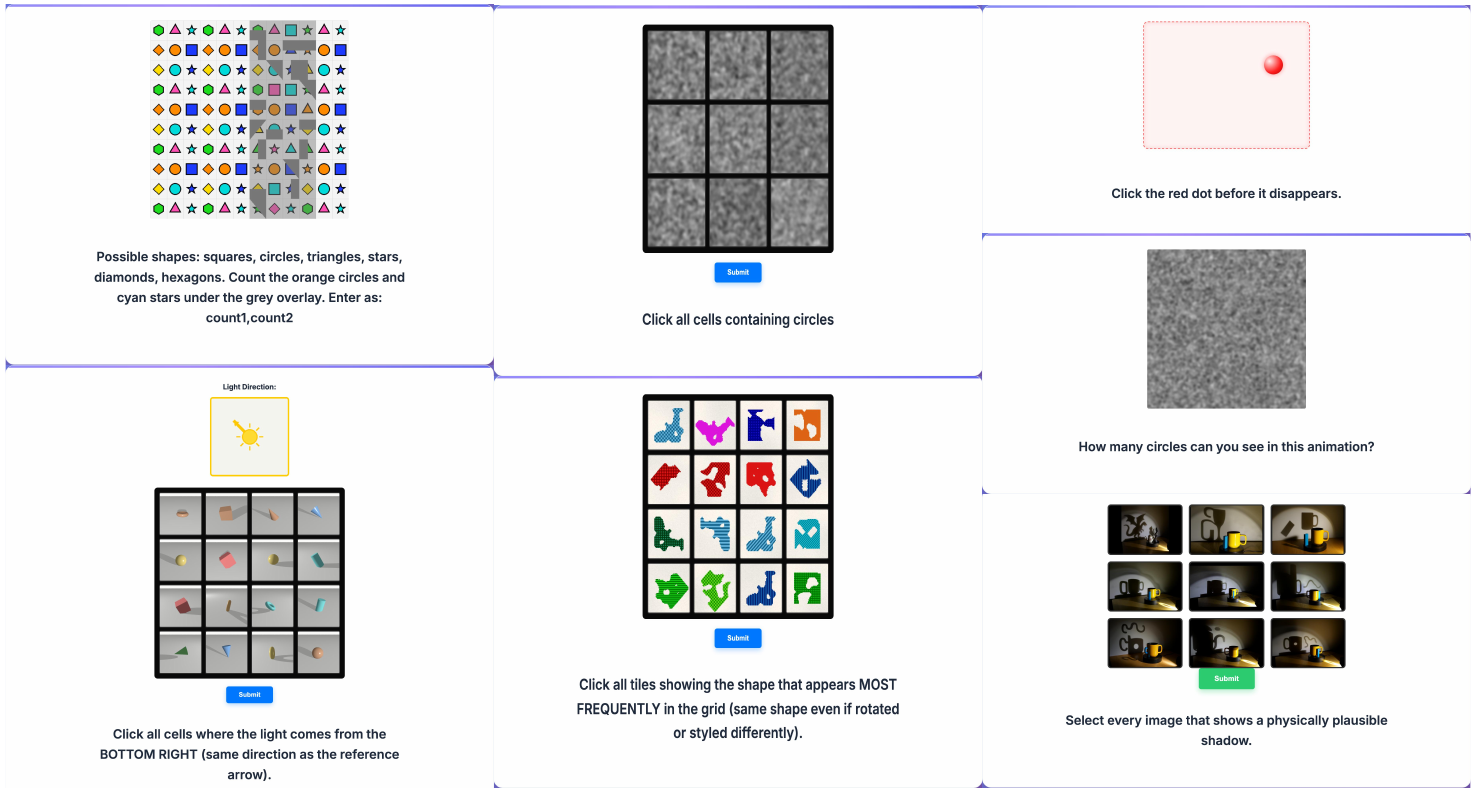}
    \vspace{-0.6em}
    \caption{\textbf{Seven concrete examples of Next-Gen CAPTCHAs.}
    Occluded Pattern Counting, Shadow Direction, Spooky Circle Grid, Rotation Match, Red Dot, Spooky Circle and Shadow Plausible.}
    \label{example2}
    \vspace{-.5em}
\end{figure*}

\begin{figure*}[t]
    \centering
    \includegraphics[width=1.\linewidth]{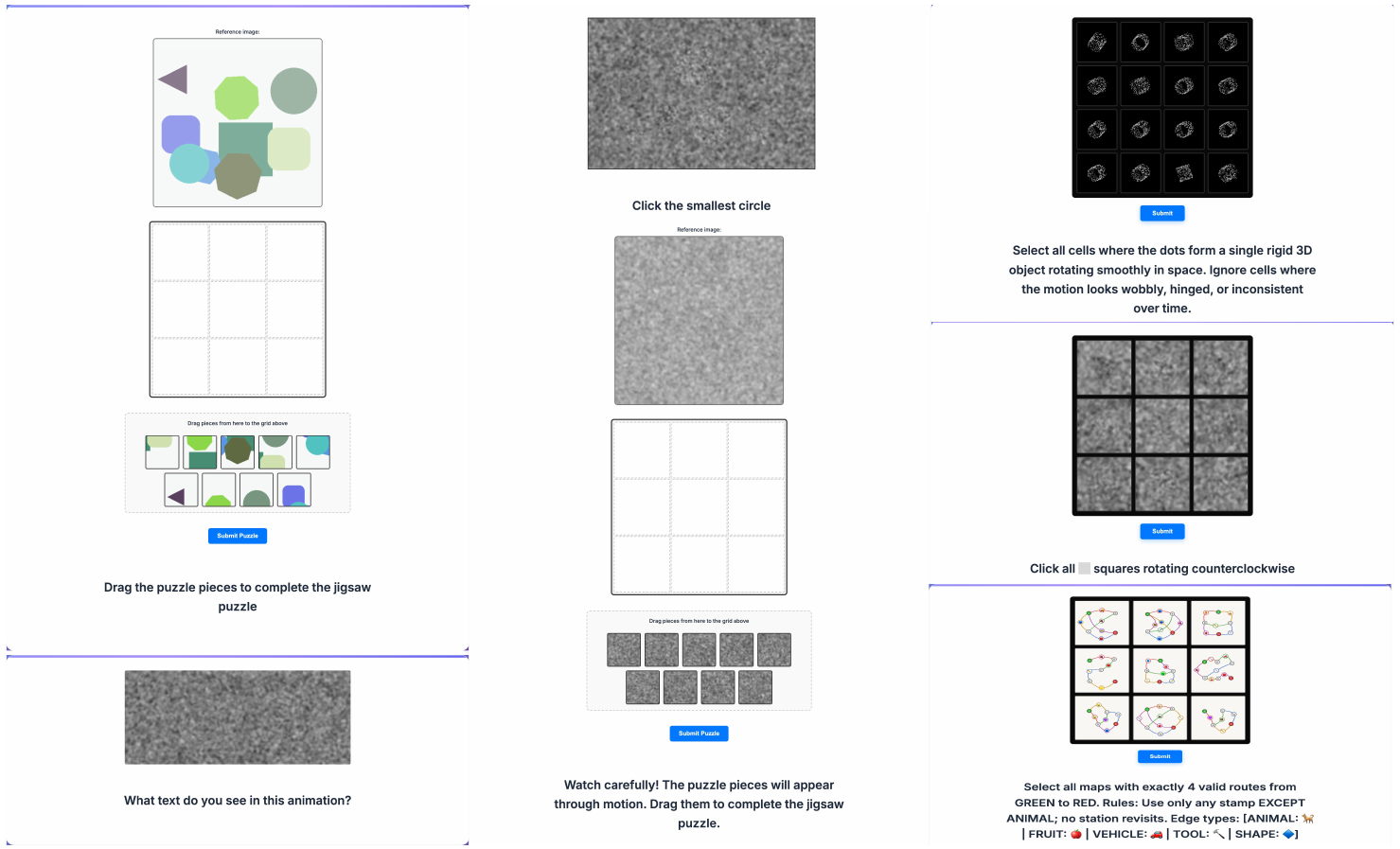}
    \vspace{-0.6em}
    \caption{\textbf{Seven concrete examples of Next-Gen CAPTCHAs.}
    Static Jigsaw, Spooky Text, Spooky Size, Spooky Jigsaw, Structure From Motion, Spooky Shape Grid and Subway Paths.}
    \label{example3}
    \vspace{-.5em}
\end{figure*}
\begin{figure*}[h]
    \centering
    \includegraphics[width=1.\linewidth]{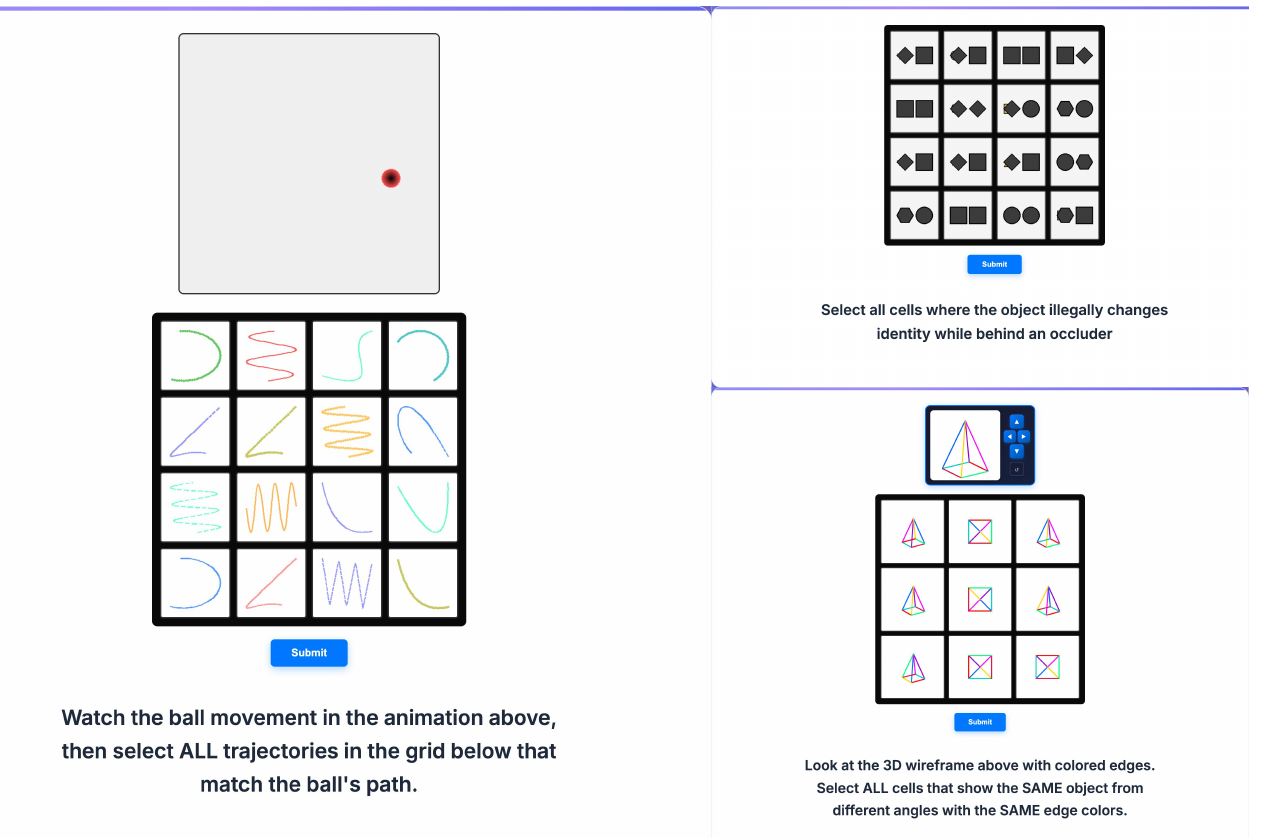}
    \vspace{-0.6em}
    \caption{\textbf{Three concrete examples of Next-Gen CAPTCHAs.} Trajectory Recovery, Temporal Object Continuity and 3D Viewpoint.}
    \label{example4}
    \vspace{-.5em}
\end{figure*}

\end{document}